\documentclass{ijuc}

\usepackage[pdftex]{graphicx}
\usepackage{amsmath,amssymb}
\usepackage{bm}
\usepackage{multirow}
\usepackage{booktabs}
\usepackage{xcolor}

\newif\ifshowchanges
\showchangesfalse 

\ifshowchanges
  \newcommand{\rev}[1]{\textcolor{red}{#1}}
\else
  \newcommand{\rev}[1]{#1}
\fi
\usepackage[paperwidth=6in, paperheight=9in, textwidth=108mm, textheight=180mm, centering]{geometry}

\begin{document}

\title{VLAGuard: A Framework for Evaluating and Mitigating Physical Attention Hijacking in Vision-Language-Action Robots within Wireless Sensor Networks}

\author{
Dongfu Yin\inst{1}\email{yindongfu@gml.ac.cn}
\and
Jinquan Zhang\inst{1,2}\email{2410815010@mails.szu.edu.cn (Corresponding author)}
}

\institute{
Guangdong Laboratory of Artificial Intelligence and Digital Economy (SZ),
Shenzhen, China
\and
Shenzhen University, Shenzhen, China
}

\def\received{}

\maketitle

\begin{abstract}
Deploying Vision-Language-Action (VLA) robots as mobile edge nodes within wireless sensor networks (WSNs) requires robust protection against physical adversarial threats. We present VLAGuard, a framework to assess and mitigate a critical vulnerability: \emph{policy-critical action-to-vision attention hijacking}. We first introduce a stress-test module, Visuomotor Attention-guided Semantic Attack (VASA), using printable patches to \rev{severely distract} the robot's action-conditioned cross-attention. To counter this, we propose Attention-Protective Fine-Tuning (APFT), a defense that stabilizes spatial-temporal attention and enforces geometric consistency with zero inference overhead. Evaluations across simulated and physical WSN-assisted smart environments demonstrate significant robustness gains. APFT reduces the OpenVLA failure rate from 100.0\% to 25.9\% in LIBERO simulations. Furthermore, across 2,000 real-world trials, APFT improves the average success rate from 23.0\% to 67.4\% under severe patch attacks. This highlights that protecting attention pathways is \rev{important for improving the robustness of} VLA-driven edge nodes in sensor networks.
\end{abstract}


\section{Introduction}
\label{sec:introduction}

Vision-Language-Action (VLA) foundation policies have become a promising paradigm for embodied intelligence by enabling robots to map RGB observations and natural-language instructions directly to control actions \cite{rt1,rt2,octo,openvla,pi0}. Supported by large-scale pretraining on internet-scale vision-language data and robot interaction data, modern VLA models show strong generalization across tasks, objects, and embodiments \cite{rt2,brohan2023openx,openvla,octo}. However, their robustness in the physical world remains insufficient for safe deployment, especially when these robots are integrated as mobile sensing nodes within broader wireless sensor networks (WSNs). \rev{This deployment trend is consistent with the emerging Internet of Humanoids (IoH) paradigm, where humanoid and embodied robots are interconnected into collaborative, intelligent networks \cite{11346994}. Related low-cost sensing interfaces for robust and dexterous human-robot interaction further highlight the importance of reliable physical sensing when robots operate in human-centered environments \cite{11378175}.}

Unlike passive vision systems, VLA robots operate in a closed perception--action loop, where localized perceptual errors can immediately induce incorrect actions and accumulate into severe long-horizon failures \cite{mft,libero,evavla}. Within a WSN context, these mobile manipulation nodes frequently depend on resource-constrained, single-view perception. Their reliable execution relies entirely on accurately grounding a sparse set of task-critical visual features---such as the precise geometry of an object's affordance or the real-time spatial relationship between the end-effector and the target. Among realistic physical threats, adversarial patches are particularly practical because they are localized, printable, and robust under real-world transformations \cite{advpatch,physical_stop_sign,eot}. Existing attacks and defenses have mainly focused on action deviation, global representation corruption, adversarial fine-tuning, or test-time purification \cite{vla_vuln,edpa_vla,advla,madry,diffpure,patchguard}. Yet, robotic manipulation often depends on this sparse task-critical evidence, suggesting that robustness depends not only on preserving \emph{what} the model represents generally, but strictly on preserving \emph{where} the policy attends during the sub-second generation of action tokens.

Despite the rapid adoption of VLA systems, the fundamental fragility of their internal perception-to-action routing remains unaddressed. When deployed in unconstrained physical environments, the assumption that a VLA model will consistently align its internal attention with task-relevant physical geometry is dangerously optimistic. Current defense mechanisms---which often rely on computationally heavy test-time purification or global adversarial fine-tuning---either introduce unacceptable latency for real-time edge control or fail to guarantee the precise spatial grounding required for physical manipulation. Therefore, there is a critical need for a framework that investigates and secures the precise neural bottleneck linking visual observation to physical action before widely deploying these embodied intelligence agents.

Motivated by this mechanistic gap, we identify a specific and \rev{severe} mechanism-level vulnerability in VLA control, termed \emph{policy-critical action-to-vision attention hijacking}. In many VLA architectures, action generation relies on a small set of latent action-query tokens that obtain task-relevant visual evidence through cross-attention \cite{rt2,openvla,pi0}. We observe that a carefully optimized physical patch can act as a dominant attention attractor, diverting action-conditioned attention away from the gripper, target object, and local affordance regions toward the patch itself. Once this grounding is corrupted, the policy loses the spatial and geometric evidence required for stable control, causing \rev{significant kinematic deviation}. Although related attention-redirection phenomena have been reported in multimodal foundation models \cite{vlm_jailbreak,visual_prompt_injection}, their consequence is significantly more severe in robotics because the resulting error propagates directly into unsafe physical behavior.

Based on this mechanistic insight, VLAGuard introduces a red-teaming module via \emph{Visuomotor Attention-guided Semantic Attack} (VASA), an Expectation-over-Transformation (EOT)-optimized printable patch that explicitly attracts policy-critical action-to-vision attention toward the patch while disrupting vision-language semantic alignment. On the blue-teaming side, VLAGuard introduces \emph{Attention-Protective Fine-Tuning} (APFT), a zero-inference-overhead defense that updates only the visual encoder in a teacher--student framework. APFT combines feature anchoring, policy-critical attention distillation, temporal attention consistency, and language-guided geometric consistency to suppress attention hijacking while preserving clean-task competence. Evaluations across simulated and physical WSN-assisted smart environments demonstrate significant robustness gains. APFT reduces the OpenVLA failure rate from 100.0\% to 25.9\% in LIBERO simulations. Furthermore, across 2,000 real-world trials, APFT improves the average success rate from 23.0\% to 67.4\% under severe patch attacks.

\textbf{Main Contributions:} The primary contributions of this work are summarized as follows:
\begin{itemize}
    \item \textbf{Comprehensive Security Framework (VLAGuard):} We identify a critical mechanism-level vulnerability in VLA control---policy-critical action-to-vision attention hijacking. To systematically address this, we propose VLAGuard, a holistic red-blue teaming framework designed to evaluate and mitigate physical adversarial threats for VLA-driven edge nodes in WSNs.
    \item \textbf{Mechanism-Aware Red-Teaming (VASA):} As the stress-test module of the framework, we introduce the Visuomotor Attention-guided Semantic Attack (VASA). It is an EOT-optimized physical patch attack that \rev{severely disrupts} embodied manipulation by maliciously dominating cross-attention, demonstrating the \rev{severe consequences} of attention hijacking.
    \item \textbf{Zero-Inference-Overhead Defense (APFT) and Massive-Scale Validation:} On the blue-teaming side, we propose Attention-Protective Fine-Tuning, a novel teacher--student robustification method that enforces spatial grounding and temporal attention consistency. We rigorously validate the entire framework across LIBERO simulations and 2,000 independent real-world trials. Results show APFT neutralizes mechanism-level attacks without adding inference latency, improving the physical success rate from 23.0\% to 67.4\% under severe conditions.
\end{itemize}

\begin{figure}[htbp!]
    \centering
    \includegraphics[width=0.95\textwidth]{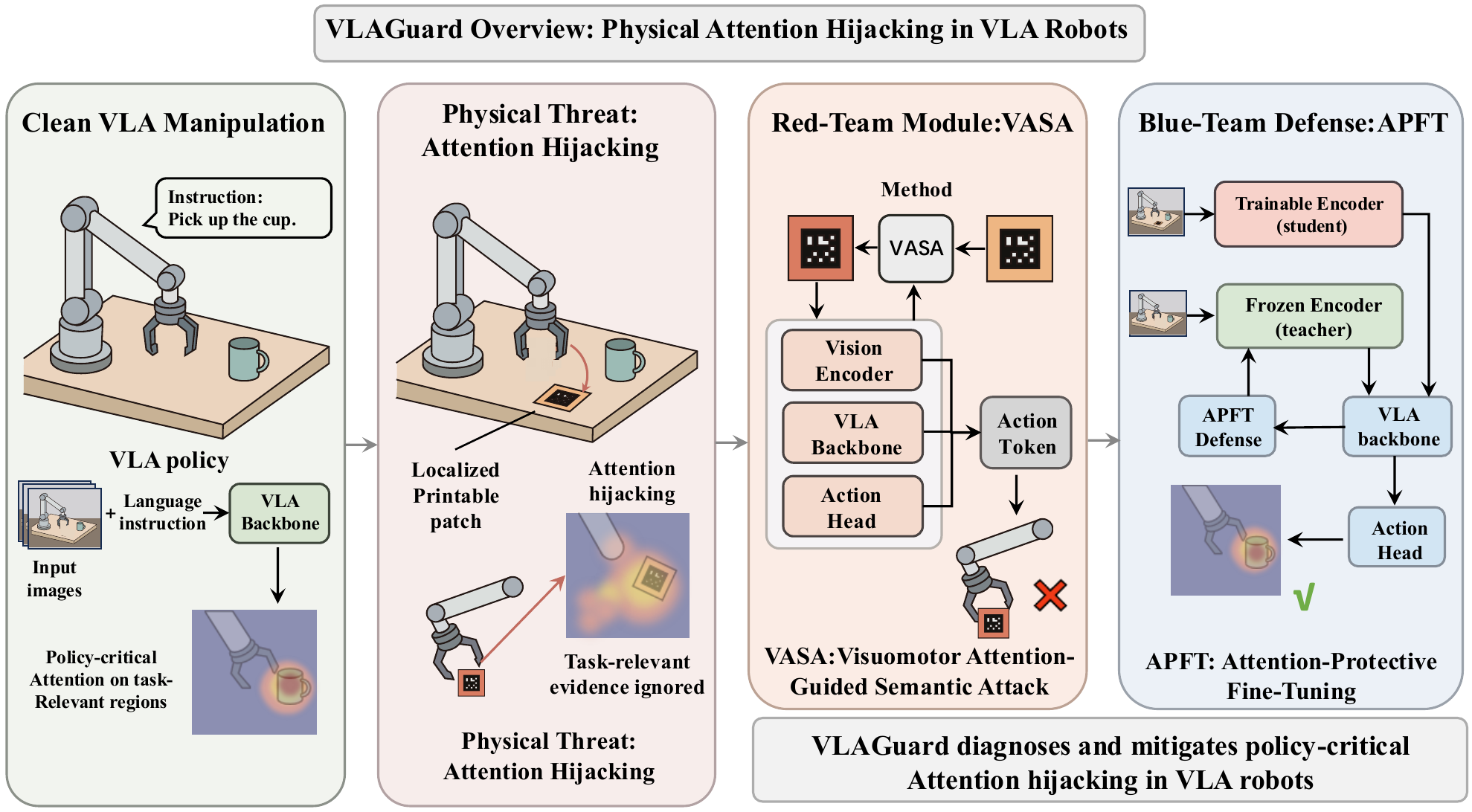}
    \caption{Overview of VLAGuard. A clean VLA policy grounds policy-critical attention on task-relevant regions for successful manipulation, whereas a localized printable patch can hijack action-to-vision attention and induce failure. VLAGuard addresses this threat with a red-team module, VASA, for mechanism-aware physical stress testing, and a blue-team module, APFT, for restoring task-relevant grounding and stabilizing the temporal evolution of policy-critical attention with zero additional inference overhead.}
    \label{fig:overview}
\end{figure}

\section{Related Work}
\label{sec:relatedwork}

\subsection{Evolution of VLA Models and Physical Deployment Challenges}
Vision-Language-Action (VLA) models have fundamentally shifted the paradigm of general-purpose robot manipulation. Systems such as RT-1 and RT-2 \cite{rt1,rt2}, OpenVLA \cite{openvla}, Octo \cite{octo}, and $\pi_0$ \cite{pi0} bypass modular pipelines by mapping multimodal observations directly to control actions. Their progress has been enabled by large-scale vision-language pretraining and heterogeneous robot datasets, as exemplified by Open X-Embodiment and related efforts \cite{brohan2023openx,openvla,octo}. \rev{Recent personalized video-chat systems further show that multimodal models can perform subject-aware temporal understanding from only one reference example \cite{Shi_2025_ICCV}, suggesting a broader trend toward adaptive perception for human-centric embodied environments.} \rev{Beyond RGB and video-based inputs, event-stream multimodal models such as EventGPT extend multimodal understanding to asynchronous event data under high-dynamic motion and challenging illumination \cite{Liu_2025_CVPR}, highlighting the importance of sensor-diverse perception for embodied agents deployed in physical environments.} Furthermore, diffusion- and flow-based policy formulations have expanded the VLA design space toward continuous control \cite{chi2023diffusionpolicy,liu2024rdt,pi0}. While large-scale pretraining has endowed these systems with unprecedented semantic generalization, their deployment as autonomous edge nodes within physical environments---such as Wireless Sensor Networks (WSNs)---imposes strict requirements on real-time reliability and precise spatial grounding. Unlike digital agents, VLA robots operate in a continuous physical space where actions irreversibly alter the environment state. Consequently, the research focus is increasingly shifting from scaling model capabilities in simulation to ensuring operational robustness in the unpredictable physical world.

\subsection{The Progression of Physical Adversarial Threats in Robotics}
As VLA models transition to physical deployment, they inherit the vulnerabilities of deep visual encoders. Physical adversarial attacks extend digital adversarial examples to realistic settings \cite{goodfellow}, particularly printable patches optimized via Expectation-over-Transformation (EOT) \cite{eot}. These pose a severe, realistic threat because they persist across varied viewpoints, lighting, printing artifacts, and sensor noise \cite{advpatch,physical_stop_sign}. In robotics, such attacks are especially catastrophic because perceptual errors propagate through sequential decision-making and accumulate into severe long-horizon task failures \cite{mft,evavla}. Recent studies demonstrate that VLA policies are vulnerable to perturbations that induce action deviation, global feature discrepancy, multimodal misalignment, or trigger-based failures \cite{vla_vuln,edpa_vla,advla,lu2025obeypatch,yan2025alignmentfails,wang2025freezevla,li2025attackvla,zhou2025badvla,zhou2025goalbackdoor,xu2025tabvla}. However, these studies predominantly analyze the system as a black box or focus on the degradation of overall scene comprehension. In embodied manipulation, a task fails not necessarily because the robot forgets what an object is, but because it loses the precise geometric coordinates required to grasp it. The exact mechanism by which localized patches dismantle the sequential control pipeline has remained underexplored.

\subsection{Limitations of Existing Defense Mechanisms}
To counter physical perturbations, contemporary research has explored two primary avenues: training-time robustification (e.g., adversarial training, adversarial fine-tuning, distillation, and parameter-efficient adaptation \cite{madry,lwf,hu2021lora,peft_survey}) and test-time purification or masking (e.g., diffusion-based denoising and patch suppression \cite{diffpure,patchguard}). Prior work has also emphasized the importance of evaluating defenses under adaptive attacks to avoid gradient obfuscation \cite{athalye2018obfuscated,carlini2017detection}. In the VLA setting, EDPA \cite{edpa_vla} provides an important baseline by adversarially fine-tuning the visual encoder to reduce global clean--patched representation discrepancy. Yet, a critical limitation persists: test-time defenses often introduce substantial inference latency, making them unviable for VLA policies that require high-frequency, closed-loop control on resource-constrained edge hardware. Conversely, existing robust fine-tuning methods typically prioritize the alignment of global multimodal representations. While effective for passive classification, global feature alignment fails to guarantee the localized, high-resolution spatial awareness required for robotic grasping and manipulation. Current defenses protect the semantic representation of the scene, but fail to protect the specific pathways used to generate physical actions.

\subsection{Attention Vulnerabilities and Our Approach}
The structural bottleneck linking generalized perception to specific actions in modern VLAs is the cross-attention mechanism. Recent findings in multimodal foundation models indicate that attention routing is inherently fragile. In large language models, compact adversarial prompts can redirect aligned generation behavior \cite{zou2023uat}. In vision-language models, adversarial patches, hidden visual instructions, and prompt-injection-style perturbations can hijack cross-modal reasoning and induce unsafe or incorrect outputs \cite{vlm_jailbreak,visual_prompt_injection}. Related observations are also emerging in robot-oriented multimodal settings \cite{jones2025robogcg}. These findings suggest that cross-attention is not merely a fusion module, but a structurally vulnerable interface. For embodied agents, this structural flaw is devastating: hijacked attention translates directly into spatial disorientation and kinematic failure. Addressing this exact progression of vulnerabilities, our work introduces the VLAGuard framework. By exposing how localized patches can maliciously dominate policy-critical attention (via the VASA attack), we move beyond global feature corruption to target the control mechanism itself. Consequently, our defense (APFT) is specifically designed to enforce spatial and temporal consistency within these attention pathways, offering a targeted, zero-inference-overhead solution that directly resolves the shortcomings of prior holistic defense strategies.

\section{Background and Threat Model}
\label{sec:background}

We briefly introduce the VLA control formulation and the physically realizable patch adversary considered in this work.

\subsection{Taxonomy of Physical Vulnerabilities in VLA Control}
\label{subsec:taxonomy}

We group physical vulnerabilities in VLA control into three levels according to where the perturbation disrupts the perception--action loop:
\begin{enumerate}
    \item \textbf{Perception-Level Disruption:} perturbing the raw visual pathway and degrading the encoder's feature extraction ability \cite{goodfellow};
    \item \textbf{Alignment-Level Disruption:} weakening the semantic binding between visual observations and language instructions \cite{edpa_vla};
    \item \textbf{Mechanism-Level Disruption (Attention Hijacking):} corrupting the cross-attention interface through which latent action tokens query task-relevant visual evidence.
\end{enumerate}
This work focuses on the third level, where localized perturbations distort policy-critical attention during action generation.

\subsection{Action Generation via Cross-Attention in VLAs}
\label{subsec:bg_vla}

A Vision-Language-Action (VLA) policy maps an RGB observation $\mathbf{x}_t \in \mathbb{R}^{H \times W \times 3}$ and a language instruction $L$ to an action $\mathbf{a}_t$:
\begin{equation}
\mathbf{a}_t = \pi(\mathbf{x}_t, L),
\label{eq:vla_mapping}
\end{equation}
where $\mathbf{a}_t$ may denote discretized actions, continuous end-effector motions, gripper commands, or hybrid outputs depending on the architecture \cite{rt2,openvla,pi0,chi2023diffusionpolicy}.

In many VLA models, the image is encoded into patch-level visual tokens and action generation relies on a small set of latent action queries, denoted by $\mathcal{Q}_{\text{act}}$, which attend to visual tokens through cross-attention:
\begin{equation}
\text{Attention}(\mathbf{Q}_{\text{act}}, \mathbf{K}_v, \mathbf{V}_v)
=
\text{Softmax}\!\left(
\frac{\mathbf{Q}_{\text{act}}\mathbf{K}_v^{T}}{\sqrt{d_k}}
\right)\mathbf{V}_v,
\label{eq:cross_attention}
\end{equation}
where $\mathbf{Q}_{\text{act}}$ is the action-query matrix, $\mathbf{K}_v$ and $\mathbf{V}_v$ are visual keys and values, and $d_k$ is the key dimensionality.

This cross-attention interface is the key bottleneck studied in this work: manipulation often depends on sparse visual evidence such as end-effector pose, object pose, and local affordance geometry, so robust control requires preserving policy-critical attention over task-relevant regions.

\subsection{Physical Threat Model}
\label{subsec:bg_threat}

We consider a physically realizable adversary that inserts a localized printable patch into the robot workspace. In practical deployments, these VLA robots often serve as intelligent, mobile visual sensor nodes within broader wireless sensor networks (WSNs). The adversarial patch aims to corrupt the local perception of these edge nodes before the sensory data can be securely processed or cross-verified, following the standard adversarial-patch setting \cite{advpatch,physical_stop_sign,eot}. Let $\boldsymbol{\delta} \in \mathbb{R}^{h \times w \times 3}$ denote the patch and let $\mathbf{m} \in \{0,1\}^{H \times W}$ denote its binary mask. The compromised observation is
\begin{equation}
\mathbf{x}'_t
=
(\mathbf{1}-\mathbf{m}) \odot \mathbf{x}_t
+
\mathbf{m} \odot \boldsymbol{\delta},
\label{eq:patch_formulation}
\end{equation}
where $\odot$ denotes the Hadamard product.

Because a physical patch must remain effective under viewpoint and environmental variation, we adopt the Expectation-over-Transformation (EOT) threat model \cite{eot}. The patch is optimized by minimizing the VASA objective under sampled physical transformations.
\begin{equation}
\boldsymbol{\delta}^{*}
=
\operatorname*{argmin}_{\boldsymbol{\delta}\in\mathcal{C}}
\;
\mathbb{E}_{\xi \sim \mathcal{T}}
\left[
\mathcal{L}_{\text{VASA}}
\Big(
\pi(\xi(\mathbf{x}'_t),L)
\Big)
\right],
\label{eq:eot_objective}
\end{equation}
where $\xi(\cdot)$ denotes a sampled physical transformation from the EOT distribution $\mathcal{T}$, $\mathcal{L}_{\text{VASA}}$ is the attack objective, and $\mathcal{C}$ denotes the printable color constraint set. Following prior work, $\mathcal{T}$ may include rotation, scaling, translation, perspective change, illumination variation, and mild sensor noise \cite{eot,physical_stop_sign}.

During evaluation, the attacker is assumed to be adaptive and white-box, with gradient access to the target policy and, when applicable, to the defense. At the same time, the perturbation remains physically constrained: it must be localized, printable, and robust under realistic image acquisition. Because VLA control is sequential, even a single erroneous action can alter later states and observations, making recovery progressively harder over a rollout. Therefore, the attacker need not dominate the full visual field; it is sufficient to persistently disrupt the alignment between action-conditioned attention and task-relevant visual evidence.

\section{The Threat: Policy-Critical Attention Hijacking}
\label{sec:threat}

As identified in Section~\ref{sec:background}, the cross-attention interface is highly vulnerable to \emph{policy-critical action-to-vision attention hijacking}. Based on this diagnosis, we introduce the Visuomotor Attention-guided Semantic Attack (VASA).

\subsection{Visuomotor Attention-guided Semantic Attack (VASA)}
\label{subsec:VASA}

Existing physical attacks on VLA systems mainly target action deviation, global feature discrepancy, or generic image-text corruption \cite{vla_vuln,edpa_vla,advla}. In contrast, VASA directly targets the pathway through which action tokens obtain task-relevant visual evidence. Its design follows a simple principle: if a localized printable patch can consistently dominate policy-critical action-to-vision attention, then the policy will lose access to the sparse spatial evidence needed for stable control.

\begin{figure}[htbp!]
    \centering
    \includegraphics[width=0.9\textwidth]{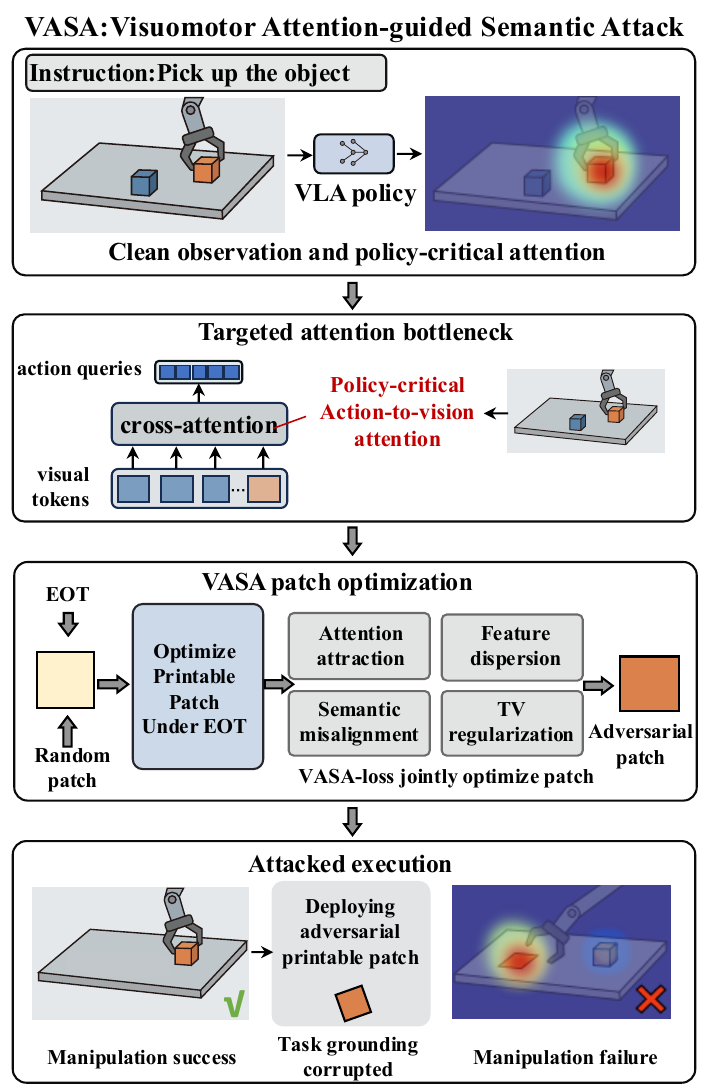}
    \caption{Pipeline of VASA. Starting from a clean observation, VASA inserts a localized printable patch and optimizes it under Expectation-over-Transformation (EOT). The attack explicitly targets the action-to-vision bottleneck by minimizing a composite objective that drives policy-critical attention toward the patch, while simultaneously maximizing feature dispersion and vision-language misalignment and enforcing physical smoothness for printability. The optimized patch redirects action-conditioned attention away from task-relevant evidence and causes manipulation failure.}
    \label{fig:vasa_pipeline}
\end{figure}

The overall VASA objective is defined as
\begin{equation}
\mathcal{L}_{\text{VASA}} =
\lambda_{\text{attn}} \mathcal{L}_{\text{attn}}
-
\lambda_{\text{disp}} \mathcal{L}_{\text{disp}}
-
\lambda_{\text{misalign}} \mathcal{L}_{\text{misalign}}
+
\lambda_{\text{tv}} \mathcal{L}_{\text{tv}},
\label{eq:VASA_total}
\end{equation}
where $\mathcal{L}_{\text{attn}}$ is the attention-guidance term, $\mathcal{L}_{\text{disp}}$ is a feature-dispersion term, $\mathcal{L}_{\text{misalign}}$ is an image-text misalignment term, and $\mathcal{L}_{\text{tv}}$ is a total-variation regularizer. Because $\mathcal{L}_{\text{disp}}$ and $\mathcal{L}_{\text{misalign}}$ are subtracted in (\ref{eq:VASA_total}), minimizing $\mathcal{L}_{\text{VASA}}$ effectively maximizes these two losses, thereby disrupting feature consistency and image-text alignment.

Among these terms, the most critical one is the attention-guidance objective. Let $K_{\text{patch}}$ denote the set of visual key tokens whose spatial support falls inside the patch mask $\mathbf{m}$. We define
\begin{equation}
\mathcal{L}_{\text{attn}}
=
-\frac{1}{|\mathcal{Q}_{\text{act}}||K_{\text{patch}}|}
\sum_{q \in \mathcal{Q}_{\text{act}}}
\sum_{k \in K_{\text{patch}}}
\bar{A}_{q,k},
\label{eq:lattn}
\end{equation}
where $\bar{A}_{q,k}$ denotes the mean cross-attention weight from action query $q$ to patch token $k$, aggregated over selected heads and late Transformer layers. Minimizing (\ref{eq:lattn}) increases the attention mass assigned to the patch region and directly drives hijacking behavior.

To destabilize semantic visual features, we adopt an InfoNCE-style feature-dispersion objective:
\begin{equation}
\mathcal{L}_{\text{disp}}
=
-\log
\frac{
\exp(\mathrm{sim}(\mathbf{z}_{\text{adv}},\mathbf{z}_{\text{clean}})/\tau_{\text{nce}})
}{
\sum_{j=1}^{B}
\exp(\mathrm{sim}(\mathbf{z}_{\text{adv}},\mathbf{z}_{\text{clean}}^{(j)})/\tau_{\text{nce}})
},
\label{eq:ldisp}
\end{equation}
where $\mathbf{z}_{\text{adv}}$ and $\mathbf{z}_{\text{clean}}$ denote the embeddings of the adversarial and clean images, $B$ is the batch size, $\tau_{\text{nce}}$ is the temperature, and $\mathrm{sim}(\cdot,\cdot)$ is cosine similarity. Since $\mathcal{L}_{\text{disp}}$ is subtracted in (\ref{eq:VASA_total}), the optimizer effectively maximizes it, which reduces the similarity between $\mathbf{z}_{\text{adv}}$ and $\mathbf{z}_{\text{clean}}$ and induces strong feature-space perturbations.

We further maximize image-text misalignment in the joint embedding space via
\begin{equation}
\mathcal{L}_{\text{misalign}}
=
\frac{1}{B}
\sum_{i=1}^{B}
\left\|
\mathrm{sim}(\mathbf{z}_{\text{adv}}^{(i)},\mathbf{z}_{\text{text}}^{(i)})
-
\mathrm{sim}(\mathbf{z}_{\text{clean}}^{(i)},\mathbf{z}_{\text{text}}^{(i)})
\right\|_{1},
\label{eq:lmisalign}
\end{equation}
where $\mathbf{z}_{\text{text}}^{(i)}$ is the embedding of the $i$-th text instruction. By maximizing this $L_1$ distance through (\ref{eq:VASA_total}), the attacker forces the image-text matching of the adversarial image to deviate significantly from the correct alignment of the clean image.

To improve physical smoothness and printability, we additionally regularize the patch with a total-variation term:
\begin{equation}
\mathcal{L}_{\text{tv}}
=
\sum_{u,v}
\left(
\left\|
\boldsymbol{\delta}_{u+1,v}-\boldsymbol{\delta}_{u,v}
\right\|_{1}
+
\left\|
\boldsymbol{\delta}_{u,v+1}-\boldsymbol{\delta}_{u,v}
\right\|_{1}
\right),
\label{eq:ltv}
\end{equation}
where $\boldsymbol{\delta}_{u,v}$ denotes the RGB value of the patch at spatial location $(u,v)$. Minimizing $\mathcal{L}_{\text{tv}}$ encourages locally smooth printable patterns and improves robustness under blur, mild viewpoint change, and camera noise.

Together, these terms yield a localized physical perturbation that acts as a strong mechanism-aware stress test for policy-critical attention in VLA systems. Because the cross-attention mass is normalized, increasing fixation on the patch necessarily reduces the attention available to task-relevant evidence. In closed-loop robotic control, this loss of grounding quickly propagates into kinematic derailment, failed grasps, and long-horizon task collapse.

\section{The Defense: Attention-Protective Fine-Tuning with Temporal Attention Consistency}
\label{sec:defense}

The analysis in Section~\ref{sec:threat} shows that physical patches disrupt not only visual features at individual timesteps, but also the spatiotemporal grounding pathway through which action-conditioned queries acquire task-relevant evidence over a rollout. In practice, this often produces an \emph{attention stickiness} failure mode: once the patch captures policy-critical action-to-vision attention, the policy tends to remain fixated on it across subsequent steps, leading to cumulative errors and long-horizon failure.

To address this issue, we extend Attention-Protective Fine-Tuning (APFT) from snapshot-level recovery to trajectory-level stabilization. In addition to preserving clean semantic features and per-step policy-critical attention, APFT explicitly constrains the temporal evolution of action-to-vision attention. The resulting defense protects both \emph{where} the policy attends and \emph{how} its attention should move during execution.

\begin{figure}[htbp!]
    \centering
    \includegraphics[width=0.9\textwidth]{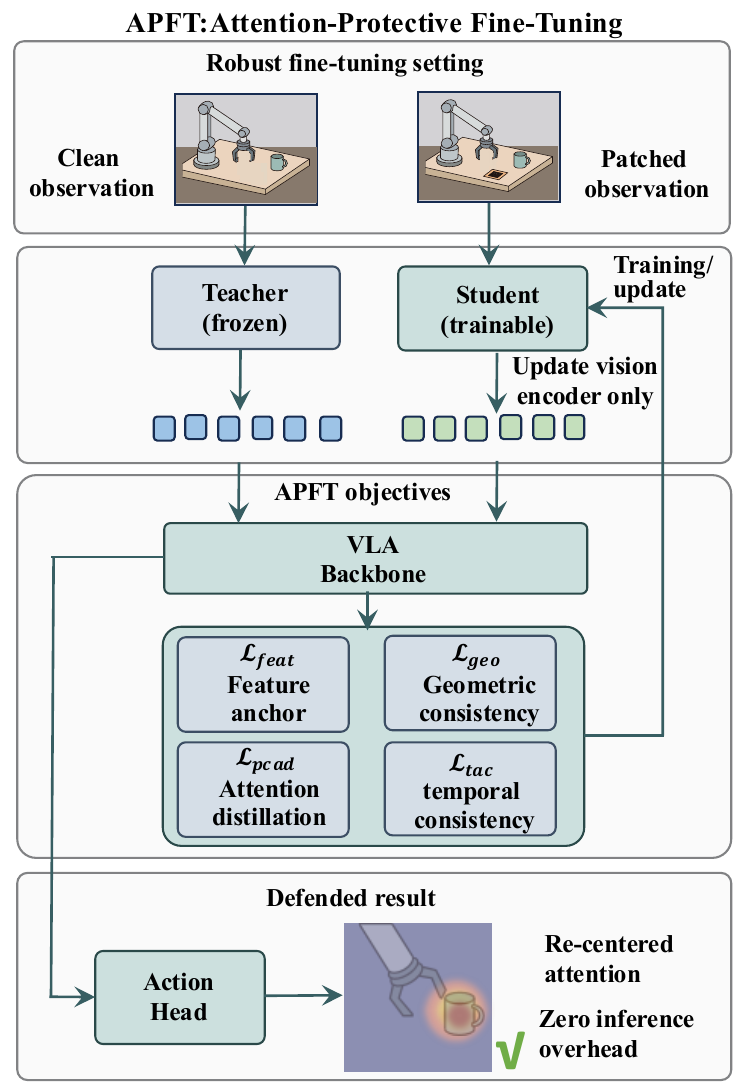}
    \caption{Pipeline of APFT. APFT adopts a teacher-student robust fine-tuning framework in which the frozen Teacher processes clean visual inputs and the trainable Student processes adversarially patched visual inputs. Robustness is injected by updating only the Student visual encoder with feature anchoring, policy-critical attention distillation, language-guided geometric consistency, and temporal attention consistency, yielding a defended VLA policy with restored spatial-temporal grounding and zero additional inference overhead.}
    \label{fig:apft_pipeline}
\end{figure}

\subsection{Teacher--Student Robust Fine-Tuning}
\label{subsec:apft_framework}

APFT is a training-time-only defense and introduces no auxiliary masking or purification module at inference time. We use a frozen pretrained VLA model as the Teacher ($T$), and initialize a trainable Student ($S$) from the same checkpoint. During robust fine-tuning, the Teacher processes clean trajectory windows and the Student processes the corresponding adversarially patched windows.

Let
\begin{equation}
\tau = (\{\mathbf{x}_t\}_{t=1}^{K}, L)
\label{eq:trajectory}
\end{equation}
denote a clean trajectory window of length $K$, and let $\tau'$ denote its adversarial counterpart generated under the VASA threat model. The Teacher receives $\mathbf{x}_t$, whereas the Student receives $\mathbf{x}'_t$. The language backbone and action head remain frozen, and only the Student visual encoder is updated. The overall APFT objective is
\begin{equation}
\mathcal{L}_{\text{APFT}}
=
\lambda_{\text{feat}}\mathcal{L}_{\text{feat}}
+
\lambda_{\text{pcad}}\mathcal{L}_{\text{pcad}}
+
\lambda_{\text{geo}}\mathcal{L}_{\text{geo}}
+
\lambda_{\text{tac}}\mathcal{L}_{\text{tac}},
\label{eq:lAPFT_total}
\end{equation}
where $\mathcal{L}_{\text{feat}}$ is a feature anchor term, $\mathcal{L}_{\text{pcad}}$ is the policy-critical attention distillation loss, $\mathcal{L}_{\text{geo}}$ is the language-guided geometric consistency loss, and $\mathcal{L}_{\text{tac}}$ is the proposed temporal attention consistency loss.

To prevent catastrophic forgetting during robust fine-tuning, we enforce the Student's patch-level semantic features to remain directionally consistent with those of the clean Teacher:
\begin{equation}
\mathcal{L}_{\text{feat}}
=
1-
\frac{1}{N}
\sum_{i=1}^{N}
\frac{
\mathbf{z}_{S}^{(i)} \cdot \mathbf{z}_{T}^{(i)}
}{
\|\mathbf{z}_{S}^{(i)}\| \, \|\mathbf{z}_{T}^{(i)}\|
},
\label{eq:lfeat}
\end{equation}
where $N$ is the number of visual patch tokens, and $\mathbf{z}_{S}^{(i)}$ and $\mathbf{z}_{T}^{(i)}$ are the $i$-th patch-level feature vectors from the Student and Teacher, respectively. This loss anchors the feature space by minimizing the cosine distance $(1-\cos\theta)$.

\subsection{Policy-Critical Attention Distillation}
\label{subsubsec:lpcad}

The core protective component of APFT is policy-critical attention distillation, which restores the Teacher's clean allocation of action-to-vision attention. Rather than aligning all attention maps, APFT focuses on the action queries $\mathcal{Q}_{\text{act}}$ that directly condition robot behavior. Let $\mathbf{P}^{T}_{t,h,q}$ and $\mathbf{P}^{S}_{t,h,q}$ denote the Teacher and Student attention distributions over visual tokens for timestep $t$, head $h$, and action query $q$. We define
\begin{equation}
\mathcal{L}_{\text{pcad}}
=
\frac{1}{KH|\mathcal{Q}_{\text{act}}|}
\sum_{t=1}^{K}\sum_{h=1}^{H}\sum_{q\in\mathcal{Q}_{\text{act}}}
D_{\text{JS}}\!\left(\mathbf{P}^{T}_{t,h,q}\,\|\,\mathbf{P}^{S}_{t,h,q}\right),
\label{eq:lpcad}
\end{equation}
where $D_{\text{JS}}(\cdot\|\cdot)$ denotes the Jensen--Shannon divergence. This term directly counteracts attention hijacking by forcing the Student to recover clean-like action-conditioned grounding at each timestep.

\subsection{Language-Guided Geometric Consistency}
\label{subsubsec:lgeo}

Localized physical patches may corrupt not only the patch region itself, but also the interpretation of nearby textures and affordance geometry. Instead of enforcing full-image alignment, APFT preserves the local structure of semantically relevant regions. For each timestep, a text-guided relevance mask is derived from the Teacher's text-to-vision attention, and cosine-normalized pairwise feature correlations are computed for both Teacher and Student features. Using the resulting relevance weights $\mathbf{M}_{t,ij}$ and correlation entries $G^{T}_{t,ij}$ and $G^{S}_{t,ij}$, the geometric consistency loss is
\begin{equation}
\mathcal{L}_{\text{geo}}
=
\frac{1}{K}
\sum_{t=1}^{K}
\frac{
\sum_{i,j}\mathbf{M}_{t,ij}\bigl(G^{S}_{t,ij}-G^{T}_{t,ij}\bigr)^2
}{
\sum_{i,j}\mathbf{M}_{t,ij}+\epsilon
},
\label{eq:lgeo}
\end{equation}
where $\epsilon$ is a small smoothing constant. This term preserves semantically relevant local structure while tolerating divergence in irrelevant regions.

\subsection{Temporal Attention Consistency}
\label{subsubsec:ltac}

While $\mathcal{L}_{\text{pcad}}$ aligns action-to-vision attention at each timestep, it does not constrain how that attention evolves over time. Under physical patch attacks, a policy may still exhibit unstable dynamics, such as prolonged patch fixation or delayed re-centering. To address this issue, we introduce Temporal Attention Consistency (TAC), which aligns the step-to-step change of the Student's action-to-vision attention with that of the Teacher. Let $\Delta\mathbf{P}^{T}_{t,h,q}$ and $\Delta\mathbf{P}^{S}_{t,h,q}$ denote the corresponding temporal attention differences from timestep $t-1$ to $t$, and let $\boldsymbol{\omega}_t$ be a Teacher-guided token-weight vector emphasizing behaviorally relevant regions. The TAC loss is
\begin{equation}
\begin{split}
\mathcal{L}_{\text{tac}}
=&\;
\frac{1}{(K-1)H|\mathcal{Q}_{\text{act}}|}
\sum_{t=2}^{K}\sum_{h=1}^{H}\sum_{q\in\mathcal{Q}_{\text{act}}} \\
&\;
\left\|
\boldsymbol{\omega}_t\odot
\bigl(\Delta\mathbf{P}^{S}_{t,h,q}-\Delta\mathbf{P}^{T}_{t,h,q}\bigr)
\right\|_2^2 ,
\end{split}
\label{eq:ltac}
\end{equation}
where $\odot$ denotes element-wise multiplication. Unlike $\mathcal{L}_{\text{pcad}}$, which enforces correct instantaneous grounding, $\mathcal{L}_{\text{tac}}$ regularizes the temporal \emph{motion} of policy-critical attention, thereby suppressing patch-induced stickiness and improving robustness in long-horizon manipulation.

\subsection{Training and Deployability}
\label{subsec:apft_training}

In each iteration, APFT samples a clean trajectory window, generates its VASA-corrupted counterpart under EOT, feeds the clean window to the frozen Teacher and the adversarial window to the Student, computes the four losses in (\ref{eq:lAPFT_total}), and updates only the Student visual encoder. Because APFT is purely a training-time defense, it introduces no extra module, no auxiliary inference branch, and no additional test-time computation. The defended policy therefore preserves the original forward pathway of the base VLA model, which is critical for real-time closed-loop control.

\section{Experimental Setup}
\label{sec:setup}

This section summarizes the evaluation protocol for the proposed mechanism--attack--defense pipeline in both simulation and real-world deployment, including the benchmark environment, base models, physical platform, baselines, and implementation details. An overview is provided in Table~\ref{tab:protocol}.

\rev{Throughout the experiments, we use \emph{Clean} to denote the no-attack condition and \emph{Original} to denote the undefended pretrained policy. We report Failure Rate (FR) in simulation because LIBERO rollouts are naturally evaluated by task failure under controlled perturbations, while Success Rate (SR) is reported in real-robot experiments because physical task completion is more intuitively measured by successful execution.}

\begin{table}[htbp!]
    \centering
    \caption{Summary of the Experimental Protocol}
    \label{tab:protocol}
    \small{
    \begin{tabular}{l p{0.65\textwidth}}
    \hline\hline
    \textbf{Parameter} & \textbf{Configuration Setting} \\
    \hline
    \textbf{Simulation} & LIBERO benchmark (Spatial, Object, Goal, Long). \\
    Episodes & 50 independent evaluation episodes per task. \\
    Metric & Failure Rate (FR, \%) $\downarrow$. \\
    \hline
    \textbf{Real-Robot} & PiPER tabletop manipulator. \\
    Physical Scale & \textbf{2,000 total trials} (5 tasks $\times$ 4 conditions $\times$ 100). \\
    Randomizations & Viewpoint, target pose, distractor placement. \\
    Lighting & Normal, dim (-30\% lux). \\
    \hline
    \textbf{Optimization} & AdamW, LR $= 2 \times 10^{-5}$, weight decay $= 0.01$. \\
    APFT Weights & $\lambda_{\text{feat}}=0.5$, $\lambda_{\text{pcad}}=1.0$, $\lambda_{\text{geo}}=0.3$, $\lambda_{\text{tac}}=0.3$, $\tau_{\text{mask}}=0.5$, $\tau_{\text{temp}}=0.5$, trajectory window length $K=4$. \\
    \hline\hline
    \end{tabular}
    }
\end{table}

\subsection{Simulation Benchmark and Base Models}
\label{subsec:setup_sim}

\textbf{Simulation benchmark:} We use the LIBERO benchmark, which provides four suites covering distinct generalization challenges: \textit{LIBERO-Spatial}, \textit{LIBERO-Object}, \textit{LIBERO-Goal}, and \textit{LIBERO-Long}. These suites probe spatial reasoning, unseen-object generalization, language-conditioned goal following, and long-horizon execution, respectively.

\textbf{Base models:} Evaluations are conducted on two representative base VLA architectures, OpenVLA and $\pi_0$. An officially fine-tuned OpenVLA variant (OpenVLA-OFT) is additionally included to assess both intra-architecture and cross-architecture transfer.

\textbf{Simulation metric:} The primary simulation metric is Failure Rate (FR, \%). For each suite, evaluations are conducted over 50 independent rollout episodes per task. Quantitative results are reported as the mean $\pm$ standard error (SE) across 3 independent runs.

\subsection{Physical Benchmark on the PiPER Platform}
\label{subsec:setup_real}

To assess sim-to-real transfer, we construct a physical benchmark on a PiPER tabletop manipulator. The platform consists of a robotic arm with a parallel-jaw gripper, a static base camera for global observation, and a wrist-mounted camera for local perception. Collectively, these visual inputs function as a localized multi-node sensor network that streams real-time environmental data to the VLA policy.

\textbf{Tasks:} Five representative manipulation tasks are used:
\begin{enumerate}
    \item \textit{Pick \& Place},
    \item \textit{Open Drawer},
    \item \textit{Sort Cube},
    \item \textit{Pour Liquid},
    \item \textit{Stack Cube}.
\end{enumerate}

\rev{A physical trial is counted as successful only when the predefined task goal is completed without human intervention: the object is grasped and placed in the target region for Pick \& Place, the drawer is opened to the required extent for Open Drawer, the cube is moved to the correct target region for Sort Cube, the liquid is transferred into the target container without major spillage for Pour Liquid, and the cubes remain stably stacked for Stack Cube.}

\textbf{Evaluation protocol:} The physical benchmark contains 2,000 total trials (5 tasks $\times$ 4 conditions $\times$ 100). The four quantitative real-world conditions are: \emph{\rev{Clean with Original policy}}, \emph{VASA (Original)}, \emph{VASA (EDPA-AF)}, and \emph{VASA (APFT)}. Across trials, we randomize target pose, distractor placement, viewing configuration, and object-to-camera distance. Evaluation order is randomized to reduce temporal confounds (such as actuator heating). The primary real-world metric is Success Rate (SR, \%).

\subsection{Threat Configurations and Baselines}
\label{subsec:setup_baselines}

\textbf{Attack settings:} We evaluate undefended VLA policies under several localized perturbation conditions: Clean, Random Patch, UADA and UPA \cite{vla_vuln}, EDPA \cite{edpa_vla}, and the proposed VASA. UADA and UPA target action-level deviation, EDPA emphasizes global representation discrepancy and image-text inconsistency, and VASA explicitly targets policy-critical attention hijacking.

\textbf{Defense baselines:} Defense evaluation compares Original, EDPA-AF \cite{edpa_vla}, and the proposed APFT. \rev{In our notation, EDPA denotes the patch attack baseline, whereas EDPA-AF denotes the corresponding adversarial fine-tuning defense trained with EDPA-style adversarial examples.} For worst-case evaluation, VASA patches attacking APFT are adaptively re-optimized with direct gradient access to the APFT-tuned visual encoder.

\subsection{Implementation Details}
\label{subsec:setup_implementation}

\rev{The APFT loss weights were set empirically based on pilot runs, with $\lambda_{\text{pcad}}=1.0$ used as the anchor because policy-critical attention recovery is the main defensive objective. The remaining weights were chosen to avoid clean-task degradation while still improving robustness under adaptive VASA attacks. This setting was then kept fixed across all reported experiments.}
\textbf{Patch constraints and EOT:} In physical deployment, multiple printable patch sizes are evaluated, including 5 $\times$ 5 cm, 8 $\times$ 8 cm, 15 $\times$ 15 cm, and 20 $\times$ 20 cm. All optimized attacks are generated under EOT with random in-plane rotation ($\theta$ uniformly distributed from $-30^\circ$ to $30^\circ$), bounded translation within 10\% of the image extent, scaling ($s$ uniformly distributed from 0.85 to 1.15).

\textbf{Optimization:} For VASA, the loss weights are set to $\lambda_{\text{attn}} = 0.8$, $\lambda_{\text{disp}} = 0.2$, and $\lambda_{\text{misalign}} = 0.5$, making attention hijacking the dominant optimization objective. For APFT, we use AdamW with learning rate $2 \times 10^{-5}$ and weight decay 0.01. Unless otherwise specified, the defense weights are $\lambda_{\text{feat}} = 0.5$, $\lambda_{\text{pcad}} = 1.0$, $\lambda_{\text{geo}} = 0.3$, and $\lambda_{\text{tac}} = 0.3$. The temporal weighting temperature is set to $\tau_{\text{temp}} = 0.5$, the text-guided masking temperature is set to $\tau_{\text{mask}} = 0.5$, and the trajectory window length is $K = 4$. Only the visual encoder is updated during robust fine-tuning, while the language backbone and action head remain frozen. All robust fine-tuning was conducted on an NVIDIA A100 GPU, and real-world physical inference was executed using an NVIDIA RTX 4090 GPU.

\section{Results I: Mechanism Validation and Simulation Evaluation}
\label{sec:results_sim}

We first examine the effectiveness and transferability of VASA, then evaluate the robustness recovery provided by APFT, followed by attention-based mechanism validation and ablation analysis in the LIBERO simulation environment.

\subsection{Attack Effectiveness and Cross-Architecture Transfer}
\label{subsec:results_attack}

Table~\ref{tab:attack_fr_models} summarizes the Failure Rate (FR) of the three VLA models across the four LIBERO suites under varying physical patch conditions. 

\begin{table}[htbp!]
    \centering
    \caption{Attack Effectiveness: Failure Rate (FR, \%) of VLA Architectures under Physical Patch Constraints}
    \label{tab:attack_fr_models}
    \resizebox{\textwidth}{!}{
    \begin{tabular}{l l c c c c c}
    \hline\hline
    \textbf{VLA Model} & \textbf{Attack Condition} & \textbf{Spatial} & \textbf{Object} & \textbf{Goal} & \textbf{Long} & \textbf{Average FR} $\uparrow$ \\
    \hline
    \multirow{4}{*}{\textbf{OpenVLA}}
    & Clean & $14.2 \pm 0.5$ & $11.6 \pm 0.4$ & $20.8 \pm 1.5$ & $46.2 \pm 2.0$ & 23.2 \\
    & Random Patch & $35.8 \pm 1.3$ & $44.6 \pm 1.2$ & $42.0 \pm 1.2$ & $75.6 \pm 2.4$ & 49.5 \\
    & EDPA \cite{edpa_vla} & $100.0 \pm 0.0$ & $100.0 \pm 0.0$ & $100.0 \pm 0.0$ & $100.0 \pm 0.0$ & 100.0 \\
    & \textbf{VASA (Ours)} & \textbf{100.0 $\pm$ 0.0} & \textbf{100.0 $\pm$ 0.0} & \textbf{100.0 $\pm$ 0.0} & \textbf{100.0 $\pm$ 0.0} & \textbf{100.0} \\
    \hline
    \multirow{4}{*}{\textbf{OpenVLA-OFT}}
    & Clean & $2.4 \pm 0.4$ & $2.6 \pm 0.3$ & $3.0 \pm 0.6$ & $4.8 \pm 0.7$ & 3.2 \\
    & Random Patch & $10.0 \pm 1.1$ & $20.4 \pm 1.5$ & $16.2 \pm 1.3$ & $32.0 \pm 2.1$ & 19.7 \\
    & EDPA \cite{edpa_vla} & $39.7 \pm 0.9$ & $52.3 \pm 0.8$ & $80.8 \pm 0.4$ & $86.4 \pm 1.9$ & 64.8 \\
    & \textbf{VASA (Ours)} & \textbf{97.2 $\pm$ 0.6} & \textbf{93.6 $\pm$ 1.2} & \textbf{100.0 $\pm$ 0.0} & \textbf{100.0 $\pm$ 0.0} & \textbf{97.7} \\
    \hline
    \multirow{4}{*}{$\bm{\pi}_0$}
    & Clean & $3.4 \pm 0.3$ & $2.0 \pm 0.2$ & $10.4 \pm 1.1$ & $42.0 \pm 1.6$ & 14.5 \\
    & Random Patch & $6.0 \pm 0.8$ & $5.2 \pm 0.5$ & $16.8 \pm 1.4$ & $48.6 \pm 1.9$ & 19.2 \\
    & EDPA \cite{edpa_vla} & $29.8 \pm 1.6$ & $39.5 \pm 1.7$ & $44.3 \pm 2.0$ & $70.7 \pm 1.6$ & 46.1 \\
    & \textbf{VASA (Ours)} & \textbf{48.8 $\pm$ 2.1} & \textbf{50.4 $\pm$ 2.4} & \textbf{70.8 $\pm$ 1.8} & \textbf{80.2 $\pm$ 1.5} & \textbf{62.6} \\
    \hline\hline
    \end{tabular}
    }
\end{table}

\begin{figure}[htbp!]
    \centering
    \includegraphics[width=0.95\textwidth]{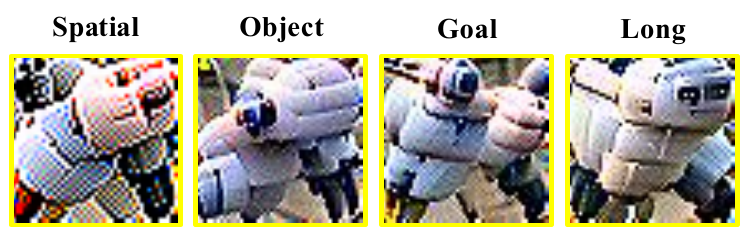}
    \caption{Visualization of the EOT-optimized VASA patches generated for the OpenVLA model across the four LIBERO benchmark suites (Spatial, Object, Goal, and Long). \rev{Each displayed patch is optimized separately for its corresponding suite, rather than being a visualization of a single shared patch.} Driven by the attention-guidance and feature-dispersion objectives, the optimization process yields distinct, uninterpretable semantic patterns that act as \rev{strong attention attractors to disrupt} the policy-critical action-to-vision cross-attention.}
    \label{fig:vasa_patches}
\end{figure}

Figure~\ref{fig:vasa_patches} visualizes the corresponding VASA patches optimized for the OpenVLA model across these four suites.

\textbf{Attack strength:} The results show that undefended VLAs are highly vulnerable to optimized physical patches. While a non-optimized random patch causes moderate degradation, the EOT-optimized VASA patch collapses the original OpenVLA policy completely, driving the FR to 100.0\% across all four suites. This indicates that once policy-critical attention is hijacked, the policy can no longer produce meaningful kinematic behavior.

\textbf{Cross-architecture transferability:} VASA also transfers strongly across models. Against OpenVLA-OFT, it raises the FR from 2.4\%--4.8\% under clean conditions to 93.6\%--100.0\%. Against $\pi_0$, a flow-matching-based architecture rather than an autoregressive one, VASA still causes severe degradation, reaching 80.2\% FR on the Long suite. This suggests that the attack targets a structural vulnerability shared by multimodal cross-attention-based policies, rather than a model-specific parameter configuration.

\subsection{Component Analysis of the Red-Teaming Module}
\label{subsec:attack_ablation}

To rigorously validate our taxonomy of physical vulnerabilities (Section~\ref{subsec:taxonomy}) and understand why VASA is so destructive, we conduct an objective-level ablation on the red-teaming module. \rev{We decompose the VASA optimization into isolated components: (1) \emph{Random} (no optimization), (2) \emph{Attn only} (optimizing only $\mathcal{L}_{\text{attn}}$ to induce mechanism-level hijacking), (3) \emph{Disp + Misalign} (optimizing $\mathcal{L}_{\text{disp}}$ and $\mathcal{L}_{\text{misalign}}$ without attention guidance), and (4) \emph{Full VASA}.}

\begin{figure}[htbp!]
    \centering
    \includegraphics[width=0.85\textwidth]{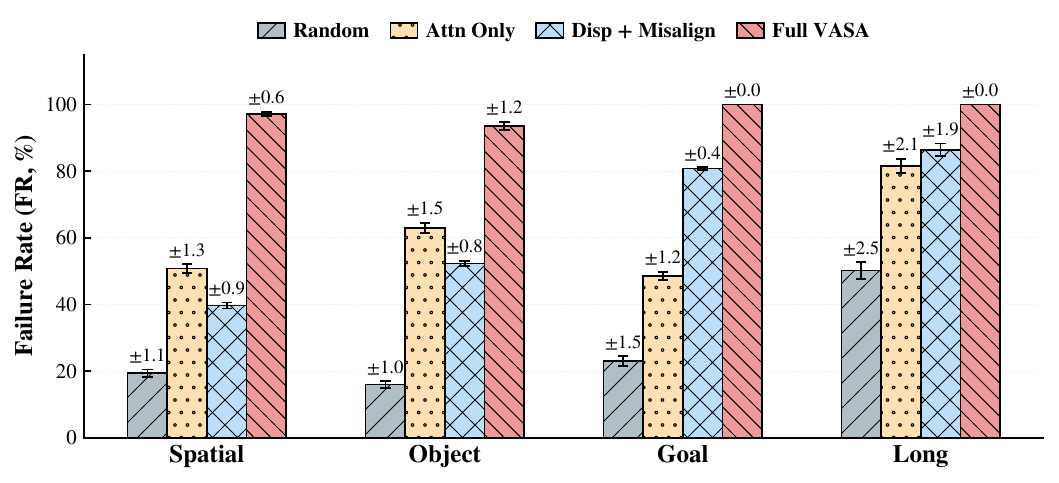}
    \caption{Ablation study of the VASA attack objectives on OpenVLA across four LIBERO suites. Solely targeting the attention bottleneck (\emph{Attn only}) significantly outperforms traditional semantic-level attacks (\emph{Disp + Misalign}), validating that mechanism-level attention hijacking is the primary driver of physical policy collapse. Error bars represent the standard error (SE) across independent evaluations.}
    \label{fig:attack_ablation}
\end{figure}

As illustrated in Figure~\ref{fig:attack_ablation}, while traditional alignment-level disruptions (\emph{Disp + Misalign}) degrade performance to some extent, explicitly targeting the cross-attention interface (\emph{Attn only}) yields a drastically higher failure rate across all suites. This empirical evidence supports our taxonomy in Section~\ref{subsec:taxonomy}, suggesting that mechanism-level attention hijacking is substantially more destructive than generic semantic disruption alone. The Full VASA objective couples these effects and produces a stringent physical stress test for policy-critical attention in VLA systems.

\subsection{Defense Robustness: APFT under Diverse Patch Attacks}
\label{subsec:results_defense}

Having established VASA as a strong stress test, we next evaluate the proposed Attention-Protective Fine-Tuning (APFT) defense. Table~\ref{tab:openvla_attack_defense} compares the undefended policy (Original), Adversarial Fine-Tuning (EDPA-AF), and APFT under multiple optimized attacks. \rev{We use OpenVLA as the primary model for the full defense and ablation analysis because it provides a representative cross-attention-based VLA architecture and exhibits the strongest vulnerability under VASA in our attack study. The cross-architecture results in Table~\ref{tab:attack_fr_models} are used to evaluate attack transferability, while the full defense sweep is concentrated on OpenVLA to keep the robust fine-tuning and ablation protocol computationally tractable.}

\begin{table}[htbp!]
    \centering
    \caption{Defense Robustness: Failure Rate (FR, \%) of OpenVLA under Various Patch Attacks}
    \label{tab:openvla_attack_defense}
    \resizebox{\textwidth}{!}{
    \begin{tabular}{l c c c c c c}
    \hline\hline
    \textbf{Defense Method} & \textbf{Attack Baseline} & \textbf{Spatial} & \textbf{Object} & \textbf{Goal} & \textbf{Long} & \textbf{Average FR} $\downarrow$ \\
    \hline
    \multirow{6}{*}{Original}
    & Clean & $14.2 \pm 0.5$ & $11.6 \pm 0.4$ & $20.8 \pm 1.5$ & $46.2 \pm 2.0$ & 23.2 \\
    & Random Patch & $35.8 \pm 1.3$ & $44.6 \pm 1.2$ & $42.0 \pm 1.2$ & $75.6 \pm 2.4$ & 49.5 \\
    & UADA \cite{vla_vuln} & $98.8 \pm 0.2$ & $94.2 \pm 1.1$ & $98.6 \pm 0.4$ & $99.8 \pm 0.2$ & 97.9 \\
    & UPA \cite{vla_vuln} & $99.0 \pm 0.1$ & $100.0 \pm 0.0$ & $96.4 \pm 0.8$ & $99.8 \pm 0.1$ & 98.8 \\
    & EDPA \cite{edpa_vla} & $100.0 \pm 0.0$ & $100.0 \pm 0.0$ & $100.0 \pm 0.0$ & $100.0 \pm 0.0$ & 100.0 \\
    & \textbf{VASA (Ours)} & 100.0 $\pm$ 0.0 & 100.0 $\pm$ 0.0 & 100.0 $\pm$ 0.0 & 100.0 $\pm$ 0.0 & 100.0 \\
    \hline
    \multirow{6}{*}{EDPA-AF \cite{edpa_vla}}
    & Clean & $17.0 \pm 0.8$ & $17.3 \pm 0.9$ & $22.8 \pm 1.2$ & $49.0 \pm 2.1$ & 26.5 \\
    & Random Patch & $19.4 \pm 1.1$ & $16.0 \pm 1.0$ & $23.0 \pm 1.5$ & $50.2 \pm 2.5$ & 27.2 \\
    & UADA \cite{vla_vuln} & $65.4 \pm 2.5$ & $43.9 \pm 2.0$ & $91.6 \pm 1.8$ & $86.7 \pm 1.5$ & 71.9 \\
    & UPA \cite{vla_vuln} & $46.6 \pm 2.1$ & $58.6 \pm 2.4$ & $68.3 \pm 2.2$ & $51.5 \pm 2.4$ & 56.3 \\
    & EDPA \cite{edpa_vla} & $39.4 \pm 1.8$ & $99.8 \pm 0.2$ & $73.9 \pm 2.5$ & $54.0 \pm 2.7$ & 66.8 \\
    & \textbf{VASA (Ours)} & $90.2 \pm 1.5$ & $100.0 \pm 0.0$ & $97.2 \pm 0.6$ & $99.0 \pm 0.5$ & 96.6 \\
    \hline
    \multirow{6}{*}{\textbf{APFT (Ours)}}
    & Clean & $14.3 \pm 0.5$ & $11.7 \pm 0.5$ & $20.9 \pm 1.2$ & $46.3 \pm 1.6$ & 23.3 \\
    & Random Patch & $14.8 \pm 0.6$ & $12.4 \pm 0.6$ & $21.8 \pm 1.4$ & $47.5 \pm 2.0$ & 24.1 \\
    & UADA \cite{vla_vuln} & $15.1 \pm 0.8$ & $12.8 \pm 0.7$ & $22.9 \pm 1.5$ & $49.2 \pm 2.2$ & 25.0 \\
    & UPA \cite{vla_vuln} & $14.9 \pm 0.7$ & $13.0 \pm 0.8$ & $22.4 \pm 1.4$ & $48.6 \pm 2.1$ & 24.7 \\
    & EDPA \cite{edpa_vla} & $15.4 \pm 0.9$ & $13.5 \pm 0.9$ & $23.8 \pm 1.6$ & $50.1 \pm 2.4$ & 25.7 \\
    & \textbf{VASA (Ours)} & \textbf{15.6 $\pm$ 0.8} & \textbf{12.8 $\pm$ 0.7} & \textbf{23.6 $\pm$ 1.5} & \textbf{51.4 $\pm$ 2.2} & \textbf{25.9} \\
    \hline\hline
    \end{tabular}
    }
\end{table}

\textbf{Robustness under adaptive attack:} Across the Spatial, Object, and Goal suites, it reduces the FR from 100.0\% to 15.6\%, 12.8\%, and 23.6\%, respectively. On average, APFT lowers the VASA-induced FR by 74.1 percentage points.

\textbf{Comparison with global alignment:} The contrast with EDPA-AF is substantial. While EDPA-AF partially improves robustness against weaker attacks, it largely collapses under adaptive VASA, yielding a 96.6\% average FR. APFT, by comparison, reduces the average FR to 25.9\%. This supports the main design principle of APFT: under severe localized physical perturbation, global feature alignment alone is insufficient, and direct protection of policy-critical attention is necessary.

\textbf{Long-horizon difficulty:} The Long suite remains the most challenging, with APFT achieving 51.4\% FR under adaptive VASA. This is consistent with the nature of long-horizon manipulation, where even small residual attention errors can accumulate over many control steps and eventually drive the policy into an unrecoverable state. Still, APFT improves the Long-suite FR by 48.6 percentage points over the undefended model.

\subsection{Mechanism Validation: Attention Re-centering}
\label{subsec:results_mechanism}

\begin{figure}[htbp!]
    \centering
    \includegraphics[width=0.95\textwidth]{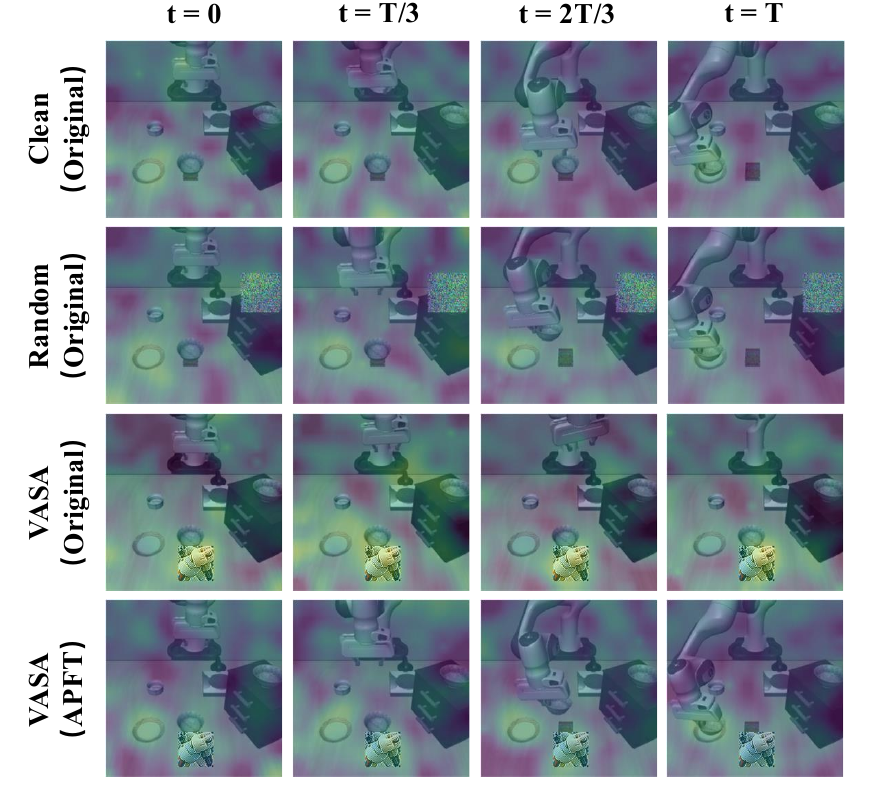}
    \caption{Temporal evolution of policy-critical action-to-vision attention in LIBERO. Under clean and random-patch conditions, the original policy maintains attention on task-relevant regions. Under VASA, attention collapses onto the adversarial patch and persists over time, whereas APFT suppresses patch fixation and restores task-relevant grounding during closed-loop execution.}
    \label{fig:attn_vis_libero}
\end{figure}

To verify that the robustness gains of APFT arise from the proposed mechanism, we visualize action-to-vision cross-attention maps from the final Transformer layers. Figure~\ref{fig:attn_vis_libero} shows their temporal evolution during a closed-loop rollout.

Under VASA, the original policy exhibits persistent \emph{patch fixation}: attention collapses onto the adversarial patch from the first timestep and remains trapped there throughout execution. As a result, the end-effector and target object are largely ignored, leading to static or divergent kinematic behavior. In contrast, under the same physical patch, the APFT-tuned policy suppresses patch fixation and re-centers attention onto task-relevant evidence, especially the gripper and target object during approach. This qualitative evidence directly links both the policy-critical attention distillation objective $\mathcal{L}_{\text{pcad}}$ and the proposed temporal attention consistency objective $\mathcal{L}_{\text{tac}}$ to the observed robustness gains. Beyond restoring correct spatial grounding, APFT also stabilizes the temporal evolution of attention, reducing prolonged patch fixation and enabling cleaner step-to-step re-centering toward task-relevant regions.

\subsection{Ablation Study of the Blue-Teaming Defense}
\label{subsec:defense_ablation}

To isolate the contribution of each defensive constraint within APFT, we evaluate OpenVLA's robustness by removing (\textit{w/o}) or selectively retaining specific loss components. In addition to the original feature, attention, and geometry terms, we explicitly analyze the role of the proposed Temporal Attention Consistency (TAC) loss.

\begin{table}[htbp!]
    \centering
    \caption{APFT Component Ablation on LIBERO. We report Failure Rate (FR, \%) under Clean, Random Patch, and adaptive VASA attacks. \rev{``Only'' means keeping only the specified loss term, whereas ``w/o'' means removing the specified term from the full APFT objective.} Lower FR indicates better robustness and utility retention.}
    \label{tab:ablation_comp_fr}
    \small{
    \begin{tabular}{l c c c}
    \hline\hline
    \multirow{2}{*}{\textbf{Method Variant}} & \multicolumn{3}{c}{\textbf{Failure Rate (FR\%$\downarrow$)}} \\
    \cmidrule(lr){2-4}
    & \textbf{Clean} & \textbf{Random} & \textbf{VASA Attack} \\
    \midrule
    only $\mathcal{L}_{\text{pcad}}$ & 33.5 $\pm$ 2.0 & 41.1 $\pm$ 2.4 & 54.4 $\pm$ 2.5 \\
    only $\mathcal{L}_{\text{geo}}$  & 27.9 $\pm$ 1.5 & 59.8 $\pm$ 1.9 & 81.6 $\pm$ 1.2 \\
    \midrule
    w/o $\mathcal{L}_{\text{feat}}$ & 31.6 $\pm$ 2.5 & 34.9 $\pm$ 2.8 & 44.4 $\pm$ 3.0 \\
    w/o $\mathcal{L}_{\text{pcad}}$ & 23.8 $\pm$ 1.1 & 44.6 $\pm$ 2.2 & 87.5 $\pm$ 1.4 \\
    w/o $\mathcal{L}_{\text{geo}}$  & 23.5 $\pm$ 1.3 & 28.5 $\pm$ 1.6 & 38.8 $\pm$ 2.1 \\
    w/o $\mathcal{L}_{\text{tac}}$  & 23.5 $\pm$ 0.9 & 24.6 $\pm$ 1.2 & 28.6 $\pm$ 1.8 \\
    \midrule
    \textbf{Full APFT} & \textbf{23.3 $\pm$ 0.6} & \textbf{24.0 $\pm$ 0.7} & \textbf{25.9 $\pm$ 1.1} \\
    \hline\hline
    \end{tabular}
    }
\end{table}

Table~\ref{tab:ablation_comp_fr} demonstrates the synergistic benefit of the full APFT objective. The most critical degradation appears when policy-critical attention distillation is removed (\textit{w/o} $\mathcal{L}_{\text{pcad}}$): the failure rate under adaptive VASA spikes to 87.5\%, indicating that spatial attention stabilization is the primary protective mechanism against hijacking. By contrast, relying solely on attention distillation (\textit{only} $\mathcal{L}_{\text{pcad}}$) also harms clean-task performance, increasing the clean FR to 33.5\%, which suggests that attention protection alone is insufficient without feature stabilization.

Furthermore, removing language-guided geometric consistency (\textit{w/o} $\mathcal{L}_{\text{geo}}$) increases the robust FR from 25.9\% to 38.8\%, showing that correct grounding must be accompanied by correct local structure interpretation. Removing temporal attention consistency (\textit{w/o} $\mathcal{L}_{\text{tac}}$) also causes a clear robustness drop, increasing the FR from 25.9\% to 28.6\%, indicating that stabilizing the temporal evolution of policy-critical attention provides additional protection beyond per-step spatial alignment. Overall, the full APFT objective achieves the best balance between clean competence and adversarial robustness.

\subsection{Physical Constraint Boundaries: Patch Scale Analysis}
\label{subsec:results_patch_scale}

To characterize the physical operating boundary of attention hijacking, we investigate how the effectiveness of the VASA attack varies with the physical size of the adversarial patch. In practical deployments, the size of a printable patch is constrained by stealth, workspace geometry, and visibility to human operators.

\begin{figure}[htbp!]
    \centering
    \includegraphics[width=0.85\textwidth]{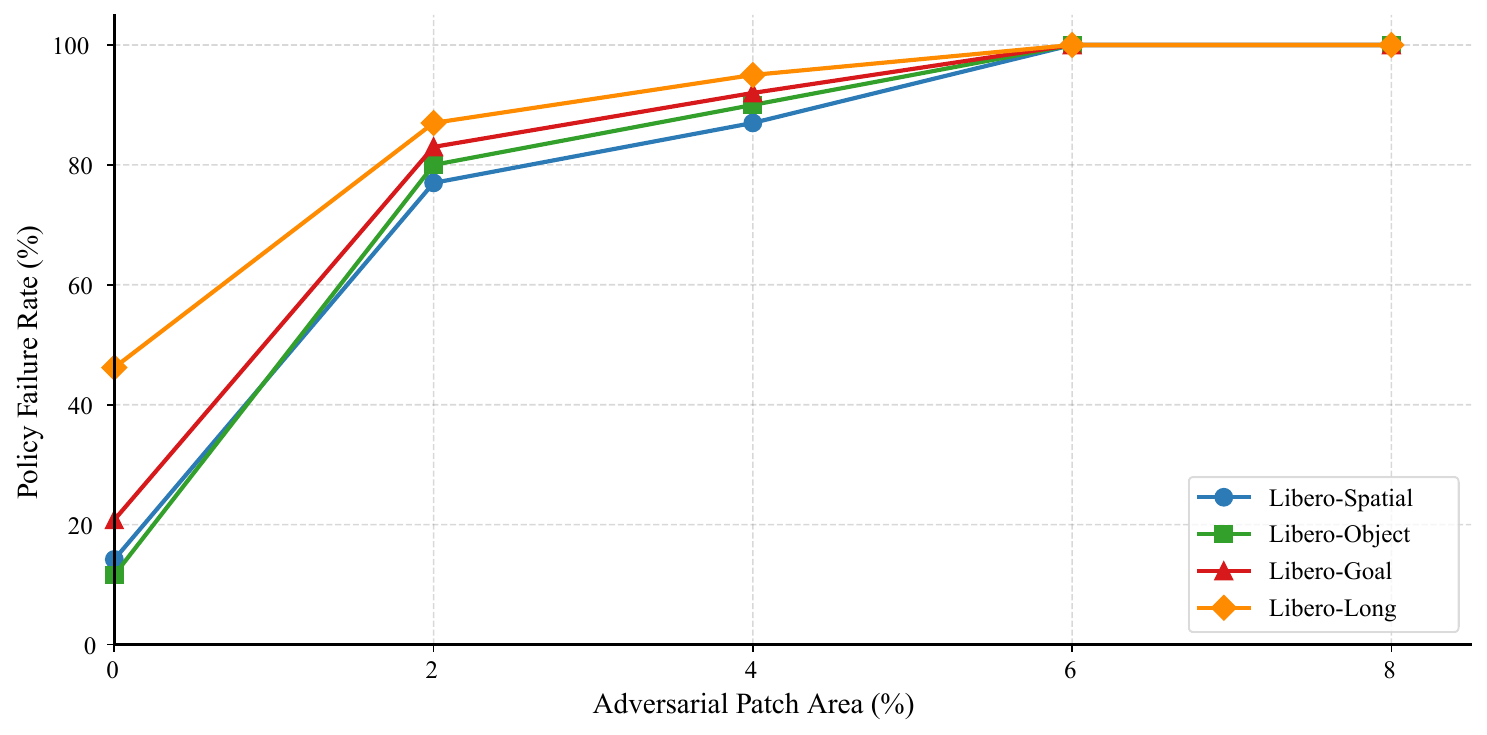}
    \caption{Impact of physical patch size on policy failure in LIBERO. As the patch area increases, the failure rate rises sharply across all four LIBERO suites, revealing a clear physical operating boundary for attention hijacking. The policy begins to degrade substantially at small patch ratios and collapses completely when the patch occupies about 5\% of the visual field, showing that severe failure does not require large-scale occlusion.}
    \label{fig:patch_scale_boundary}
\end{figure}

As illustrated in Figure~\ref{fig:patch_scale_boundary}, we evaluate the failure rate while varying the patch area from 0 to 8\% of the total field of view. The results reveal a clear threshold effect: the undefended policy experiences a rapid collapse once the patch reaches 2\% (70\%--80\% failure rate), and climbs to 80\%--90\% at a 3\% area. At a 5\% footprint, the policy is completely derailed across all suites (100\% failure). This confirms that physical attention hijacking does not require massive occlusion to be lethal.

\section{Results II: Massive-Scale Real-Robot Deployment}
\label{sec:results_real}

\begin{table}[htbp!]
  \centering
  \caption{Massive-Scale Real-Robot Deployment: Success Rate (SR, \%) Across 2,000 Independent Physical Trials. ($\Delta$ denotes absolute improvement of APFT over EDPA-AF).}
  \label{tab:real_robot}
  \resizebox{\textwidth}{!}{
  \begin{tabular}{l c c c c c}
  \hline\hline
  \multirow{2}{*}{\textbf{Task}} & \textbf{Reference} & \multicolumn{3}{c}{\textbf{Under Physical VASA Attack}} & \textbf{Improv.} \\
  \cline{3-5}
  & \textbf{Clean} & \textbf{Original} & \textbf{EDPA-AF} & \textbf{APFT (Ours)} & \textbf{($\Delta$)} \\
  \hline
  Pick \& Place & 79.0 & 28.0 & 50.0 & \textbf{76.0} & \textbf{+26.0} \\
  Open Drawer   & 75.0 & 25.0 & 45.0 & \textbf{70.0} & \textbf{+25.0} \\
  Sort Cube     & 70.0 & 24.0 & 43.0 & \textbf{66.0} & \textbf{+23.0} \\
  Pour Liquid   & 68.0 & 24.0 & 36.0 & \textbf{64.0} & \textbf{+28.0} \\
  Stack Cube    & 66.0 & 14.0 & 29.0 & \textbf{61.0} & \textbf{+32.0} \\
  \hline
  \textbf{Average SR $\uparrow$} & \textbf{71.6} & \textbf{23.0} & \textbf{40.6} & \textbf{67.4} & \textbf{+26.8} \\
  \hline\hline
  \end{tabular}
  }
\end{table}

While simulation provides a controlled setting for isolating mechanisms, the ultimate validation of a robotic security framework lies in its physical realizability. Real-world deployment introduces substantial unmodeled variability, including sensor noise, illumination changes, background distractors, and execution imprecision. In this section, we report results from a large-scale physical evaluation comprising 2,000 trials on the PiPER tabletop manipulator, bridging the sim-to-real gap for VLA robustness.

\subsection{Quantitative Recovery across Five Manipulation Tasks}
\label{subsec:real_quantitative}

Table~\ref{tab:real_robot} summarizes the success rates (SR) across five physically executed manipulation tasks. The results reveal both the severity of physical patch attacks and the effectiveness of APFT under real-world conditions.

\textbf{Physical impact of VASA:} Under clean conditions, the original policy achieves an average success rate of 71.6\%. Under the printed VASA patch, however, this drops sharply to 23.0\%, confirming that the attention hijacking mechanism identified in Section~\ref{sec:threat} persists in the physical world. The effect is especially severe on precision-heavy tasks such as \textit{Stack Cube}, where performance falls from 66.0\% to 14.0\%.

\textbf{Comparison with EDPA-AF:} EDPA-AF partially restores performance, reaching an average SR of 40.6\%, but remains limited on contact-rich and precise tasks. For example, its SR on \textit{Pour Liquid} is only 36.0\%. This result is consistent with the simulation findings: global feature alignment alone is insufficient under severe physical perturbation.

\begin{figure}[htbp!]
    \centering
    \includegraphics[width=0.95\textwidth]{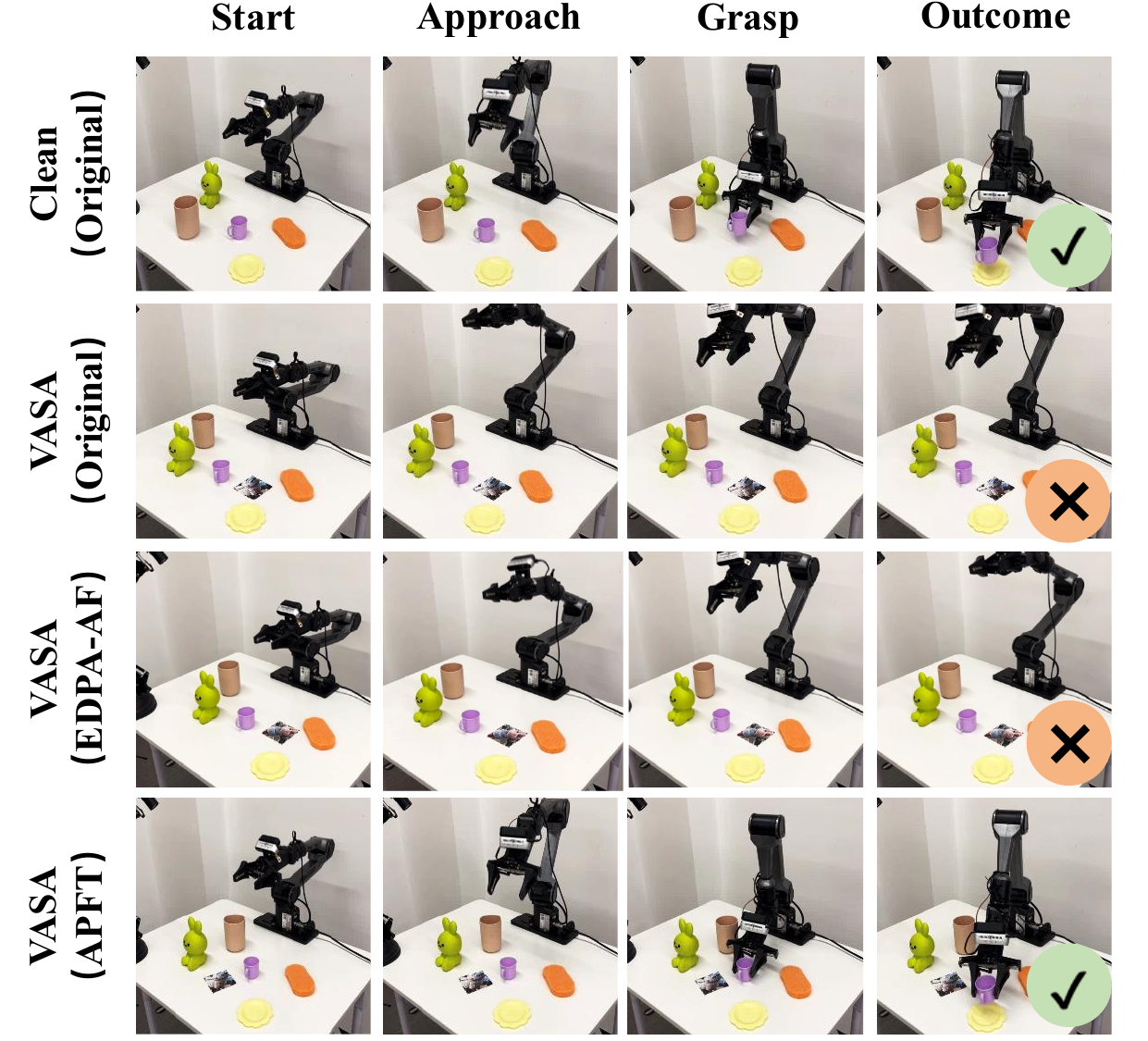}
    \caption{Representative real-world manipulation rollouts on the PiPER tabletop manipulator under clean and physical VASA attack conditions. From left to right, columns show the task progression from \emph{Start} to \emph{Approach}, \emph{Grasp}, and \emph{Outcome}. From top to bottom, rows correspond to \emph{\rev{Clean with Original policy}}, \emph{VASA (Original)},\emph{VASA (EDPA-AF)}, and \emph{VASA (APFT)}. The original policy succeeds under clean conditions but fails under the printed physical VASA patch due to corrupted task-relevant grounding. EDPA-AF provides only limited recovery, whereas APFT restores successful execution under the same attack, demonstrating improved real-world robustness against attention hijacking.}
    \label{fig:real_robot_platform}
\end{figure}

\textbf{Recovery with APFT:} The full APFT framework (including temporal consistency) consistently improves robustness across all five tasks, achieving an impressive 67.4\% average SR under severe physical VASA patches. On \textit{Pick \& Place}, APFT restores performance to 76.0\%, approaching the 79.0\% clean baseline. Even on the most demanding high-precision \textit{Stack Cube} task, APFT reaches 61.0\%, outperforming the EDPA-AF baseline by 32.0 percentage points. 

Figure~\ref{fig:real_robot_platform} visualizes representative real-world manipulation rollouts under clean conditions and under physical VASA attacks for the Original, EDPA-AF, and APFT policies, further illustrating how APFT suppresses patch fixation and restores successful execution.

Despite this substantial recovery, the policy still encounters failures in extreme corner cases, such as when the target object is completely occluded by the robotic arm during execution. These cases highlight an important limitation: although APFT mitigates algorithmic attention hijacking, it cannot overcome the physical limits of single-view RGB perception under full occlusion or severe sensor saturation. Addressing these boundaries will likely require multi-view sensing, additional modalities, or temporal memory.

\section{Discussion}
\label{sec:discussion}

\subsection{Implications and the Robustness Trade-off}
\label{subsec:implications_tradeoff}
While scaling data and parameters has substantially improved VLA generalization, our findings indicate that scaling alone cannot resolve the structural fragility of the cross-attention mechanism. A persistent challenge in adversarial defense is the ``robustness tax''---the tendency for robust fine-tuning to degrade a model's performance on clean data or unseen tasks. Our empirical results suggest that by constraining only the visual encoder and utilizing a teacher-student distillation approach, APFT effectively preserves the semantic capabilities of the frozen language backbone. However, as adaptive attackers evolve---potentially employing semantic patches that imitate language-relevant objects or targeting early visual features---future research must rigorously evaluate how anchoring visual features to defend against local patches might affect the VLA's ability to generalize to entirely novel environments or highly unconventional object geometries not present in the teacher's original distribution. Future VLA systems will likely require native attention stabilization together with defense-in-depth strategies that combine attention protection, token pruning, and stronger robustness guarantees.

\subsection{System-Level Viability for Embodied Edge Computing}
\label{subsec:system_viability}

\rev{Compared with standard adversarial training, APFT does not aim to learn generic perturbation invariance over the entire image. Instead, it selectively protects the action-conditioned attention pathways that directly influence robot control, which is especially important for manipulation tasks where small spatial grounding errors can accumulate over long horizons. Compared with patch-masking or test-time purification defenses, APFT introduces no detector, masking module, or denoising branch during deployment. This avoids additional per-step latency and reduces the risk that repeated test-time processing will disrupt closed-loop temporal consistency.}

For VLA models deployed as mobile nodes within smart environments, computational efficiency is as critical as algorithmic robustness.A significant advantage of the APFT framework is its zero-inference-overhead characteristic. Because all protective optimizations---including feature anchoring and temporal attention consistency---are strictly enforced during training, the deployed policy retains its original forward-pass architecture. This allows APFT to be seamlessly integrated into computationally constrained edge devices without introducing the latency associated with diffusion-based purification or auxiliary detection modules, successfully meeting the stringent real-time requirements of closed-loop physical control.

\subsection{Physical Boundaries and Sensory Limitations}
\label{subsec:physical_boundaries}
Although APFT effectively mitigates algorithmic attention hijacking by enforcing spatial-temporal grounding, it is fundamentally bounded by the absolute limits of single-view RGB perception. As demonstrated in our real-world failure cases, severe physical scenarios---such as complete obliteration of object affordances by robotic occlusion or extreme sensor saturation---represent hard boundaries for software-only defenses. Mitigating these physical-layer failures requires moving beyond purely algorithmic robustification. Overcoming these hardware-level sensory limits will likely require multi-modal sensor fusion (e.g., integrating depth and tactile feedback) to ensure spatial awareness is maintained even when the primary visual pathway is heavily compromised.

\subsection{Moving Toward Multi-Node Collaborative Defense in WSNs}
\label{subsec:collaborative_defense}
Currently, VLAGuard secures the perception--action loop of isolated VLA agents. However, as these robots become deeply integrated into broader Wireless Sensor Networks (WSNs) and smart infrastructures, isolated node defense will be insufficient. When a VLA robot's localized attention is maliciously hijacked, the system should ideally cross-verify its intended actions against environmental observations provided by static overhead cameras or peer robots within the WSN. \rev{Future frameworks should explore distributed consensus algorithms that fuse multi-view and cross-modal network data, inspired by recent cross-modal fusion designs for 3D perception \cite{10943919}, to detect internal attention anomalies in real-time, elevating embodied security from single-agent robustification to system-wide network resilience.}

\section{Conclusion}
\label{sec:conclusion}

In this article, we identified and addressed \emph{policy-critical action-to-vision attention hijacking}, a severe physical vulnerability that compromises VLA robots operating as intelligent mobile nodes within wireless sensor networks (WSNs). Rather than relying on computationally expensive test-time purification, our proposed VLAGuard framework employs Attention-Protective Fine-Tuning (APFT) to natively harden the perception--action loop with zero additional inference latency. As validated by extensive real-world and simulated deployments, stabilizing the spatiotemporal attention of individual robotic nodes effectively mitigates localized adversarial patch attacks (VASA) and restores reliable physical execution. 

Moving forward, as mobile manipulators become deeply integrated into distributed IoT and smart environments, ensuring the robustness of isolated edge nodes will not be sufficient. Future work will explore collaborative defense mechanisms, leveraging multi-node sensor fusion and distributed consensus algorithms within the broader wireless sensor network to cross-verify visual evidence, thereby ensuring system-level fault tolerance and secure physical interactions across the entire network ecosystem.

\section*{Acknowledgment}
\rev{This work was supported in part by the Shenzhen Science and Technology Program under Grant KJZD20240903104400001.}

\bibliography{ijuc}

@article{rt1,
  title={Rt-1: Robotics transformer for real-world control at scale},
  author={Brohan, Anthony and Brown, Noah and Carbajal, Justice and Chebotar, Yevgen and Dabis, Joseph and Finn, Chelsea and Gopalakrishnan, Keerthana and Hausman, Karol and Herzog, Alex and Hsu, Jasmine and others},
  journal={arXiv preprint arXiv:2212.06817},
  year={2022}
}

@inproceedings{rt2,
  title={Rt-2: Vision-language-action models transfer web knowledge to robotic control},
  author={Zitkovich, Brianna and Yu, Tianhe and Xu, Sichun and Xu, Peng and Xiao, Ted and Xia, Fei and Wu, Jialin and Wohlhart, Paul and Welker, Stefan and Wahid, Ayzaan and others},
  booktitle={Conference on Robot Learning},
  pages={2165--2183},
  year={2023},
  organization={PMLR}
}

@article{openvla,
  title={Openvla: An open-source vision-language-action model},
  author={Kim, Moo Jin and Pertsch, Karl and Karamcheti, Siddharth and Xiao, Ted and Balakrishna, Ashwin and Nair, Suraj and Rafailov, Rafael and Foster, Ethan and Lam, Grace and Sanketi, Pannag and others},
  journal={arXiv preprint arXiv:2406.09246},
  year={2024}
}

@article{octo,
  title={Octo: An open-source generalist robot policy},
  author={Team, Octo Model and Ghosh, Dibya and Walke, Homer and Pertsch, Karl and Black, Kevin and Mees, Oier and Dasari, Sudeep and Hejna, Joey and Kreiman, Tobias and Xu, Charles and others},
  journal={arXiv preprint arXiv:2405.12213},
  year={2024}
}

@article{pi0,
  title={{$\pi_0$}: A Vision-Language-Action Flow Model for General Robot Control},
  author={Black, Kevin and Brown, Noah and Driess, Danny and Esmail, Adnan and Equi, Michael and Finn, Chelsea and Fusai, Niccolo and Groom, Lachy and Hausman, Karol and Ichter, Brian and Jakubczak, Szymon and Jones, Tim and Ke, Liyiming and Levine, Sergey and Li-Bell, Adrian and Mothukuri, Mohith and Nair, Suraj and Pertsch, Karl and Shi, Lucy Xiaoyang and Tanner, James and Vuong, Quan and Walling, Anna and Wang, Haohuan and Zhilinsky, Ury},
  journal={arXiv preprint arXiv:2410.24164},
  year={2024}
}

@article{mft,
  title={Manipulation facing threats: Evaluating physical vulnerabilities in end-to-end vision language action models},
  author={Cheng, Hao and Xiao, Erjia and Wang, Yichi and Yu, Chengyuan and Sun, Mengshu and Zhang, Qiang and Cao, Jiahang and Guo, Yijie and Liu, Ning and Xu, Kaidi and others},
  journal={arXiv preprint arXiv:2409.13174},
  year={2024}
}

@article{advpatch,
  title={Adversarial patch},
  author={Brown, Tom B and Man{\'e}, Dandelion and Roy, Aurko and Abadi, Mart{\'\i}n and Gilmer, Justin},
  journal={arXiv preprint arXiv:1712.09665},
  year={2017}
}

@inproceedings{eot,
  title={Synthesizing robust adversarial examples},
  author={Athalye, Anish and Engstrom, Logan and Ilyas, Andrew and Kwok, Kevin},
  booktitle={International conference on machine learning},
  pages={284--293},
  year={2018},
  organization={PMLR}
}

@inproceedings{vla_vuln,
  title={Exploring the adversarial vulnerabilities of vision-language-action models in robotics},
  author={Wang, Taowen and Han, Cheng and Liang, James and Yang, Wenhao and Liu, Dongfang and Zhang, Luna Xinyu and Wang, Qifan and Luo, Jiebo and Tang, Ruixiang},
  booktitle={Proceedings of the IEEE/CVF International Conference on Computer Vision},
  pages={6948--6958},
  year={2025}
}

@article{edpa_vla,
  title={Model-agnostic adversarial attack and defense for vision-language-action models},
  author={Xu, Haochuan and Koh, Yun Sing and Huang, Shuhuai and Zhou, Zirun and Wang, Di and Sakuma, Jun and Zhang, Jingfeng},
  journal={arXiv preprint arXiv:2510.13237},
  year={2025}
}

@article{goodfellow,
  title={Explaining and harnessing adversarial examples},
  author={Goodfellow, Ian J and Shlens, Jonathon and Szegedy, Christian},
  journal={arXiv preprint arXiv:1412.6572},
  year={2014}
}

@INPROCEEDINGS{physical_stop_sign,
  author={Eykholt, Kevin and Evtimov, Ivan and Fernandes, Earlence and Li, Bo and Rahmati, Amir and Xiao, Chaowei and Prakash, Atul and Kohno, Tadayoshi and Song, Dawn},
  booktitle={2018 IEEE/CVF Conference on Computer Vision and Pattern Recognition}, 
  title={Robust Physical-World Attacks on Deep Learning Visual Classification}, 
  year={2018},
  volume={},
  number={},
  pages={1625-1634},
  doi={10.1109/CVPR.2018.00175}}

@article{evavla,
  title={Eva-VLA: Evaluating Vision-Language-Action Models' Robustness Under Real-World Physical Variations},
  author={Liu, Hanqing and Ruan, Shouwei and Long, Jiahuan and Wu, Junqi and Hou, Jiacheng and Tang, Huili and Jiang, Tingsong and Zhou, Weien and Yao, Wen},
  journal={arXiv preprint arXiv:2509.18953},
  year={2025}
}

@article{advla,
  title={Attention-Guided Patch-Wise Sparse Adversarial Attacks on Vision-Language-Action Models},
  author={Zhang, Naifu and Tao, Wei and Xiao, Xi and Sun, Qianpu and Zheng, Yuxin and Mo, Wentao and Wang, Peiqiang and Zhang, Nan},
  journal={arXiv preprint arXiv:2511.21663},
  year={2025}
}

@article{madry,
  title={Towards deep learning models resistant to adversarial attacks},
  author={Madry, Aleksander and Makelov, Aleksandar and Schmidt, Ludwig and Tsipras, Dimitris and Vladu, Adrian},
  journal={arXiv preprint arXiv:1706.06083},
  year={2017}
}

@article{diffpure,
  title={Diffusion models for adversarial purification},
  author={Nie, Weili and Guo, Brandon and Huang, Yujia and Xiao, Chaowei and Vahdat, Arash and Anandkumar, Anima},
  journal={arXiv preprint arXiv:2205.07460},
  year={2022}
}

@inproceedings{patchguard,
  title={$\{$PatchGuard$\}$: A provably robust defense against adversarial patches via small receptive fields and masking},
  author={Xiang, Chong and Bhagoji, Arjun Nitin and Sehwag, Vikash and Mittal, Prateek},
  booktitle={30th USENIX Security Symposium (USENIX Security 21)},
  pages={2237--2254},
  year={2021}
}

@article{peft_survey,
  title={Parameter-efficient fine-tuning for large models: A comprehensive survey},
  author={Han, Zeyu and Gao, Chao and Liu, Jinyang and Zhang, Jeff and Zhang, Sai Qian},
  journal={arXiv preprint arXiv:2403.14608},
  year={2024}
}

@article{lwf,
  title={Learning without forgetting},
  author={Li, Zhizhong and Hoiem, Derek},
  journal={IEEE transactions on pattern analysis and machine intelligence},
  volume={40},
  number={12},
  pages={2935--2947},
  year={2017},
  publisher={IEEE}
}

@article{vlm_jailbreak,
  title={Jailbreak vision language models via bi-modal adversarial prompt},
  author={Ying, Zonghao and Liu, Aishan and Zhang, Tianyuan and Yu, Zhengmin and Liang, Siyuan and Liu, Xianglong and Tao, Dacheng},
  journal={IEEE Transactions on Information Forensics and Security},
  year={2025},
  publisher={IEEE}
}

@article{visual_prompt_injection,
  title={Empirical analysis of large vision-language models against goal hijacking via visual prompt injection},
  author={Kimura, Subaru and Tanaka, Ryota and Miyawaki, Shumpei and Suzuki, Jun and Sakaguchi, Keisuke},
  journal={arXiv preprint arXiv:2408.03554},
  year={2024}
}

@article{libero,
  title={Libero: Benchmarking knowledge transfer for lifelong robot learning},
  author={Liu, Bo and Zhu, Yifeng and Gao, Chongkai and Feng, Yihao and Liu, Qiang and Zhu, Yuke and Stone, Peter},
  journal={Advances in Neural Information Processing Systems},
  volume={36},
  pages={44776--44791},
  year={2023}
}

@inproceedings{brohan2023openx,
  title={Open x-embodiment: Robotic learning datasets and rt-x models: Open x-embodiment collaboration 0},
  author={O’Neill, Abby and Rehman, Abdul and Maddukuri, Abhiram and Gupta, Abhishek and Padalkar, Abhishek and Lee, Abraham and Pooley, Acorn and Gupta, Agrim and Mandlekar, Ajay and Jain, Ajinkya and others},
  booktitle={2024 IEEE International Conference on Robotics and Automation (ICRA)},
  pages={6892--6903},
  year={2024},
  organization={IEEE}
}

@article{chi2023diffusionpolicy,
  title={Diffusion policy: Visuomotor policy learning via action diffusion},
  author={Chi, Cheng and Xu, Zhenjia and Feng, Siyuan and Cousineau, Eric and Du, Yilun and Burchfiel, Benjamin and Tedrake, Russ and Song, Shuran},
  journal={The International Journal of Robotics Research},
  volume={44},
  number={10-11},
  pages={1684--1704},
  year={2025},
  publisher={Sage Publications Sage UK: London, England}
}

@article{liu2024rdt,
  title={Rdt-1b: a diffusion foundation model for bimanual manipulation},
  author={Liu, Songming and Wu, Lingxuan and Li, Bangguo and Tan, Hengkai and Chen, Huayu and Wang, Zhengyi and Xu, Ke and Su, Hang and Zhu, Jun},
  journal={arXiv preprint arXiv:2410.07864},
  year={2024}
}

@inproceedings{athalye2018obfuscated,
  title={Obfuscated gradients give a false sense of security: Circumventing defenses to adversarial examples},
  author={Athalye, Anish and Carlini, Nicholas and Wagner, David},
  booktitle={International conference on machine learning},
  pages={274--283},
  year={2018},
  organization={PMLR}
}

@inproceedings{carlini2017detection,
  title={Adversarial examples are not easily detected: Bypassing ten detection methods},
  author={Carlini, Nicholas and Wagner, David},
  booktitle={Proceedings of the 10th ACM workshop on artificial intelligence and security},
  pages={3--14},
  year={2017}
}

@article{jones2025robogcg,
  title={Adversarial attacks on robotic vision language action models},
  author={Jones, Eliot Krzysztof and Robey, Alexander and Zou, Andy and Ravichandran, Zachary and Pappas, George J and Hassani, Hamed and Fredrikson, Matt and Kolter, J Zico},
  journal={arXiv preprint arXiv:2506.03350},
  year={2025}
}

@article{lu2025obeypatch,
  title={When Robots Obey the Patch: Universal Transferable Patch Attacks on Vision-Language-Action Models},
  author={Lu, Hui and Yu, Yi and Yang, Yiming and Yi, Chenyu and Zhang, Qixin and Shen, Bingquan and Kot, Alex C and Jiang, Xudong},
  journal={arXiv preprint arXiv:2511.21192},
  year={2025}
}

@article{yan2025alignmentfails,
  title={When alignment fails: Multimodal adversarial attacks on vision-language-action models},
  author={Yan, Yuping and Xie, Yuhan and Zhang, Yixin and Lyu, Lingjuan and Wang, Handing and Jin, Yaochu},
  journal={arXiv preprint arXiv:2511.16203},
  year={2025}
}

@article{wang2025freezevla,
  title={Freezevla: Action-freezing attacks against vision-language-action models},
  author={Wang, Xin and Li, Jie and Weng, Zejia and Wang, Yixu and Gao, Yifeng and Pang, Tianyu and Du, Chao and Teng, Yan and Wang, Yingchun and Wu, Zuxuan and others},
  journal={arXiv preprint arXiv:2509.19870},
  year={2025}
}

@article{li2025attackvla,
  title={AttackVLA: Benchmarking Adversarial and Backdoor Attacks on Vision-Language-Action Models},
  author={Li, Jiayu and Zhao, Yunhan and Zheng, Xiang and Xu, Zonghuan and Li, Yige and Ma, Xingjun and Jiang, Yu-Gang},
  journal={arXiv preprint arXiv:2511.12149},
  year={2025}
}

@article{zhou2025badvla,
  title={Badvla: Towards backdoor attacks on vision-language-action models via objective-decoupled optimization},
  author={Zhou, Xueyang and Tie, Guiyao and Zhang, Guowen and Wang, Hechang and Zhou, Pan and Sun, Lichao},
  journal={arXiv preprint arXiv:2505.16640},
  year={2025}
}

@article{zhou2025goalbackdoor,
  title={Goal-oriented backdoor attack against vision-language-action models via physical objects},
  author={Zhou, Zirun and Xiao, Zhengyang and Xu, Haochuan and Sun, Jing and Wang, Di and Zhang, Jingfeng},
  journal={arXiv preprint arXiv:2510.09269},
  year={2025}
}

@article{xu2025tabvla,
  title={TabVLA: Targeted Backdoor Attacks on Vision-Language-Action Models},
  author={Xu, Zonghuan and Zheng, Xiang and Ma, Xingjun and Jiang, Yu-Gang},
  journal={arXiv preprint arXiv:2510.10932},
  year={2025}
}

@article{zou2023uat,
  title={Universal and transferable adversarial attacks on aligned language models},
  author={Zou, Andy and Wang, Zifan and Carlini, Nicholas and Nasr, Milad and Kolter, J Zico and Fredrikson, Matt},
  journal={arXiv preprint arXiv:2307.15043},
  year={2023}
}

@article{hu2021lora,
  title={Lora: Low-rank adaptation of large language models.},
  author={Hu, Edward J and Shen, Yelong and Wallis, Phillip and Allen-Zhu, Zeyuan and Li, Yuanzhi and Wang, Shean and Wang, Liang and Chen, Weizhu and others},
  journal={Iclr},
  volume={1},
  number={2},
  pages={3},
  year={2022}
}

@ARTICLE{11346994,
  author={Yu, Angela W. and Nayak, Amiya},
  journal={IEEE Internet of Things Journal}, 
  title={The Internet of Humanoids: A Survey of Technologies, Applications, and Challenges}, 
  year={2026},
  volume={13},
  number={6},
  pages={10498-10521},
  }

@InProceedings{Shi_2025_ICCV,
    author    = {Shi, Yufei and Yan, Weilong and Xu, Gang and Li, Yumeng and Chen, Yucheng and Li, Zhenxi and Yu, Fei and Li, Ming and Yeo, Si Yong},
    title     = {PVChat: Personalized Video Chat with One-Shot Learning},
    booktitle = {Proceedings of the IEEE/CVF International Conference on Computer Vision (ICCV)},
    month     = {October},
    year      = {2025},
    pages     = {23321--23331}
}

@InProceedings{Liu_2025_CVPR,
    author    = {Liu, Shaoyu and Li, Jianing and Zhao, Guanghui and Zhang, Yunjian and Meng, Xin and Yu, Fei Richard and Ji, Xiangyang and Li, Ming},
    title     = {EventGPT: Event Stream Understanding with Multimodal Large Language Models},
    booktitle = {Proceedings of the Computer Vision and Pattern Recognition Conference (CVPR)},
    month     = {June},
    year      = {2025},
    pages     = {29139--29149}
}

@INPROCEEDINGS{11378175,
  author={Yin, Dongfu and Yang, Run and Xie, Lei and Yu, F. Richard and Chen, Ji and Dai, Bing},
  booktitle={2025 IEEE International Conference on Robotics and Biomimetics (ROBIO)}, 
  title={A Low-Cost Sensing Glove for Robust and Dexterous Human-Robot Interaction}, 
  year={2025},
  volume={},
  number={},
  pages={2093-2098},
  doi={10.1109/ROBIO66223.2025.11378175}}

@INPROCEEDINGS{10943919,
  author={Cai, Haojie and Yin, Dongfu and Yu, Fei Richard and Xiong, SiTing},
  booktitle={2025 IEEE/CVF Winter Conference on Applications of Computer Vision (WACV)}, 
  title={DSTR: Dual Scenes Transformer for Cross-Modal Fusion in 3D Object Detection}, 
  year={2025},
  volume={},
  number={},
  pages={3064-3073},
  doi={10.1109/WACV61041.2025.00303}}

\end{document}